\documentclass[journal]{IEEEtran}
\usepackage{graphicx}
\usepackage{comment}
\usepackage{amsmath,amssymb} 
\usepackage{color}

\usepackage{footnote}
\usepackage{floatrow}
\usepackage{dirtytalk}
\usepackage{boldline}
\usepackage{multirow}
\usepackage{rotating}

\usepackage{subcaption}
\usepackage{color,soul}
\usepackage{multirow}
\usepackage{dirtytalk}
\usepackage{xcolor}
\usepackage{tabularx,booktabs}
\newcolumntype{Y}{>{\centering\arraybackslash}X}

\newcommand{\ie}{{\it i.e. }}
\newcommand{\etal}{{\it et al. }}
\newcommand{\eg}{{\it e.g. }}

\newcommand{\etc}{{\it etc.}}

 \usepackage{subcaption}
 \usepackage{subfloat}

 \usepackage{dirtytalk}
 \usepackage{multirow}
 \usepackage{boldline}
 \usepackage{adjustbox}
 \usepackage{balance}
 \usepackage{rotating}
 \usepackage{balance}
 \newcommand{\bfsection}[1]{\vspace*{0.1cm}\noindent\textbf{#1.}}

\usepackage[pagebackref=true,breaklinks=true,letterpaper=true,colorlinks,bookmarks=false]{hyperref}

\begin{document}
%
\title{Disentangled Representation Learning for Controllable Person Image Generation}

%
%
%

	\author{{Wenju~Xu}, ~Chengjiang~Long, Yongwei~Nie
		and~Guanghui~Wang,~\IEEEmembership{Senior Member,~IEEE}
		\thanks{W. Xu is with AMAZON, Palo Alto, CA 94301 USA (e-mail: xuwenju@amazon.com).}
		\thanks{C. Long is with the META Reality Labs, Burlingame, CA, USA. Email: cjfykx@gmail.com.}
		\thanks{Y. Nie is with the South China University of Technology, Guangzhou, Guangdong, China, Email: nieyongwei@scut.edu.cn.}
		\thanks{G. Wang is with the Department of Computer Science, Toronto Metropolitan University, 350 Victoria St, Toronto, ON M5B 2K3. Email: wangcs@ryerson.ca.}.
}

%
%

\markboth{}%
{Shell \MakeLowercase{\textit{et al.}}: Bare Demo of IEEEtran.cls for IEEE Journals}
%



\maketitle

\begin{abstract}
In this paper, we propose a novel framework named DRL-CPG to learn disentangled latent representation for controllable person image generation, which can produce realistic person images with desired poses and human attributes (\eg pose, head, upper clothes, and pants) provided by various source persons. Unlike the existing works leveraging the semantic masks to obtain the representation of each component, we propose to generate disentangled latent code via a novel attribute encoder with transformers trained in a manner of curriculum learning from a relatively easy step to a gradually hard one. A random component mask-agnostic strategy is introduced to randomly remove component masks from the person segmentation masks, which aims at increasing the difficulty of training and promoting the transformer encoder to recognize the underlying boundaries between each component. This enables the model to transfer both the shape and texture of the components. Furthermore, we propose a novel attribute decoder network to integrate multi-level attributes (\eg the structure feature and the attribute representation) with well-designed Dual Adaptive Denormalization (DAD) residual blocks. Extensive experiments strongly demonstrate that the proposed approach is able to transfer both the texture and shape of different human parts and yield realistic results. To our knowledge, we are the first to learn disentangled latent representations with transformers for person image generation.
\end{abstract}


%
\IEEEpeerreviewmaketitle

\section{Introduction}
Deep generative adversarial network \cite{goodfellow2014generative,huang2020fx} has recently drawn increasing attention due to its impressive performance in image/video synthesis \cite{xu2021drb,xu2023learning}, which shows great potential in dealing with MultiMedia applications \cite{zhai2023feature,xu2016direct,yu2021monte}. For instance, exploring synthesized features has been proven to be an effective way to improve the performance of deep neural networks \cite{xu2019adversarially,yang2021adaptive,XU2019195}. Manipulating facial images \cite{karaouglu2021self,zhang2021disentangled, XU2019570} and translating human faces into anime \cite{li2021anigan,xu2021domain} have become popular in social media applications. More recently, researchers have attempted to synthesize human images that can be controlled by user inputs  \cite{hu20213dbodynet,tang2021total}. We can imagine that using synthesized digital humans for broadcasting, advertising, and educating will be promising. However, this is still an open problem that needs more endeavor and will significantly impact multimedia society. 

\begin{figure}[t]
 \begin{center}
 \includegraphics[width=0.86\linewidth]{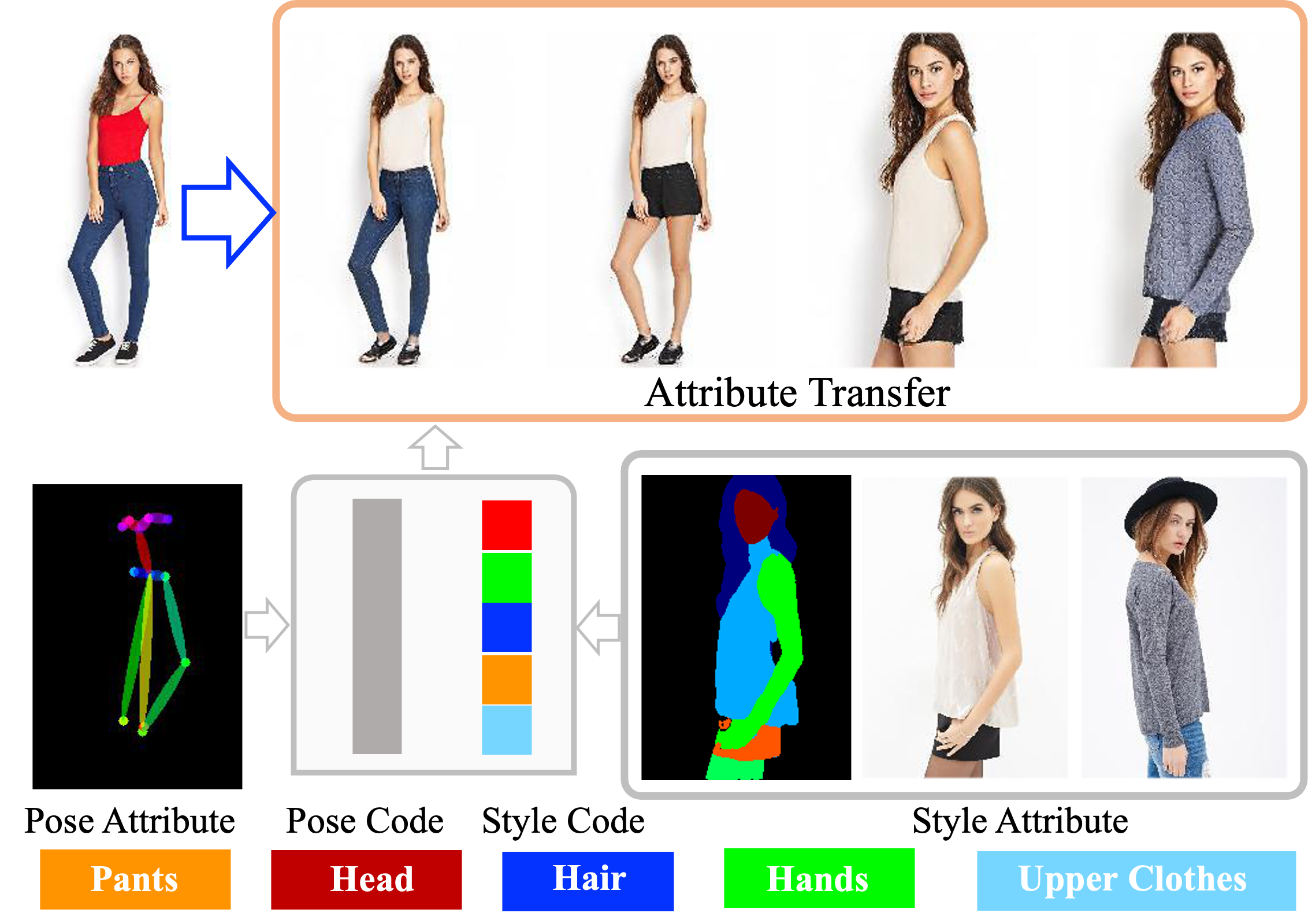}
 \end{center}
 \caption{Given a target pose and source person images with semantic masks, the goal of this paper is to design a unified approach for controllable person image generation and attribute transfer.
}
 \label{fig:introduction}
 \end{figure}

Controllable person image generation aims to synthesize a person image conditioned on the given pose and attributes at the component-level from source person images with the corresponding semantic masks, preserving attributes like person identity, cloth color, cloth texture, background \etc, as shown in Figure~\ref{fig:introduction}. 
This topic has attracted great attention due to its potentially wide applications in movie composition, image editing, person re-identification, virtual clothes try-on, and so on.


ADGAN \cite{men2020controllable} was proposed as the first work for controllable person attribute editing based on the semantic mask to separate each component. Although it achieves success in controllable image editing, the synthesized images are not realistic. In particular, separation in image level does not guarantee the disentanglement of the encoded attributes. Moreover, editing person attributes by simply replacing the entangled semantic representations tends to create artifacts or unrealistic results. 

\begin{figure*}[t]
\begin{center}
\includegraphics[width=1.\linewidth]{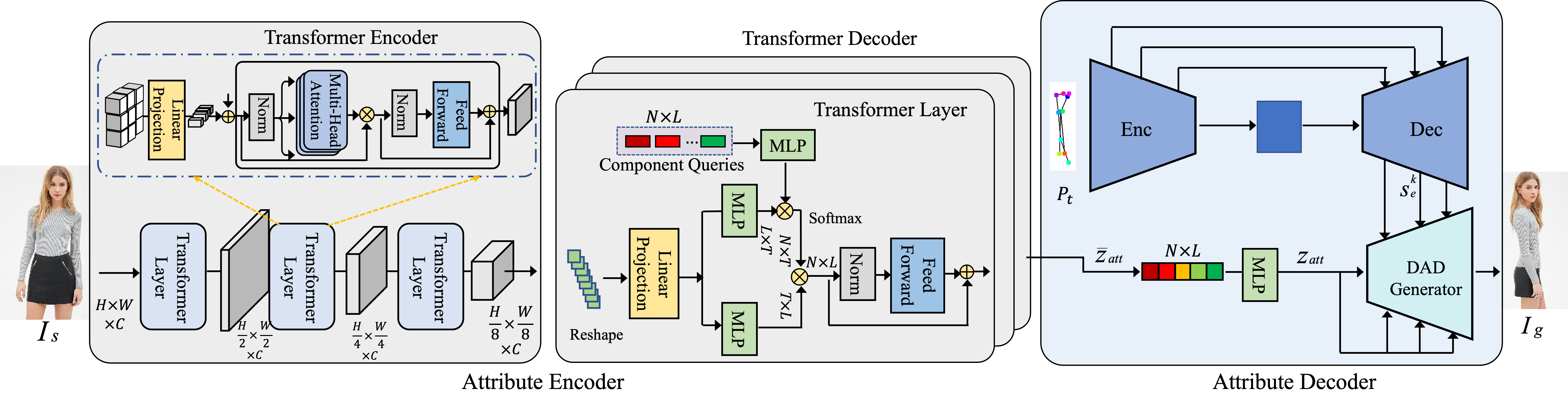}
\end{center}
\caption{The framework of our proposed DRL-CPG. It consists of an attribute encoder with transformers and an attribute decoder for pose and style transfer. The attribute encoder is trained with curriculum learning starting from a relatively easy step gradually to a complex complete mask-agnostic step, which can encode a person into latent attribute space robustly based on semantic masks. At the testing stage, the learned encoder can generate meaningful component attribute representations of a source person image. 
With the latent component attributes extracted, the attribute decoder with dually adaptive denormalization (DAD) ResBlocks integrates multiple attributes with the given pose to generate the desired image and achieve pose and attribute transfer. 
}
\label{fig:overview}
\end{figure*}

To solve the above issues of previous work, we propose a novel and unified framework, termed as DRL-CPG, for controllable person synthetic image generation. As shown in Figure~\ref{fig:overview}, the framework consists of two major parts, \ie an attribute encoder with transformers learned to generate disentangled representation and an attribute decoder that integrates the structure features and attribute representations for controllable person image generation. 
In contrast to ADGAN~\cite{men2020controllable} which encodes each component into latent code directly, we introduce transformers~\cite{vaswani2017attention} in the attribute encoder to generate an intermediate representation set for each component and select the corresponding one as the final representation of this component. 
Note that a global receptive field via self-attention provided in the transformer encoder is essential for the disentangled representation of a person's image as small components often share similar textures and context with their surroundings. Then the transformer decoder integrates learnable component queries with the transformer encoder output to generate the attribute latent representation for the subsequent attribute decoder to conduct attribute transfer.

To enable an efficient and robust way for attribute editing, we introduce a random component mask-agnostic strategy, {\em i.e.}, randomly removing several component masks from the entire person mask. The more component masks are removed, the more difficult to distinguish or recognize. Obviously, this treatment is to increase the difficulty for our model to learn how to encode the parts without masks to separate each other into disentangled semantic representations. It requires our model to robustly recognize the underlying boundaries between different components and be able to transfer both the shape and texture of components. 

Motivated by human's easy-to-hard learning process,  we adopt the curriculum learning strategy~\cite{bengio2009curriculum} and start from a relatively easy step, then gradually increase the difficulty levels to the complete mask-agnostic step. Unlike NTED~\cite{ren2022neural} relied on explicitly learned correlation matrices for feature extraction and distribution, we observe that the generated latent attribute representations are well clustered with the component mask-agnostic strategy. This easy-to-hard learning process not only enables our model to recognize the component boundary but also provides a way to preserve semantic completeness when distributing the extracted neural textures to different target poses. The learned attribute encoder can generate disentangled attribute representations for components in the person image. This significantly increases the flexibility of our model compared to PISE \cite{zhang2021pise} which requires to prediction of a parsing mask and injects texture code into different parts of the predicted mask for person generation. Without requiring predicted masks, our model avoids the issues, \eg, holes and artifacts in the generations raised by misclassified regions in the predicted masks. Our model is also able to adaptively transfer the shape of the components to fit the source person, which is not possible for mask prediction based methods without sufficient guidance to reshape the predicted masks.


Regarding our attribute decoder, it can adaptively integrate the structure feature and the attribute representation for person image generation. Note that the pose map only provides the structural connection between different joints, while it does not contain any structural information within the local region, making it difficult to synthesize rich local structures. Inspired by SPADE \cite{park2019semantic} and
AdaIN \cite{huang2017arbitrary,dumoulin2016learned}, we design Dual Adaptive Denormalization (DAD) residual blocks to explore the rich structure information for attribute transfer to 
ensure high-quality person image generation.

In summary, the novelty of our proposed DRL-CPG is mainly reflected in a novel encoder with transformers and a curriculum learning with a random mask-agnostic strategy to enforce the encoder explored to learn better representation from hard examples. This strategy creates a challenging learning task that requires holistically understanding each component region given the guidance of semantic mask and transferring the learned understanding to encode components without masks as guidance. Thus it ensures our model robustly recognizes the underlying boundaries between different components and is able to localize each component. As a result, our model learning disentangled representations of attended regions works for both pose transfer and attribute transfer with well-preserved texture details and consistent component shapes. 
Extensive experimental evaluations strongly demonstrate that our proposed DRL-CPG yields more realistic results that are more faithful to the inputs than other state-of-the-art methods in both pose transfer and component attribute transfer.

\vspace{-6pt} 

\begin{figure*}
\centering     
\includegraphics[width=0.74\linewidth]{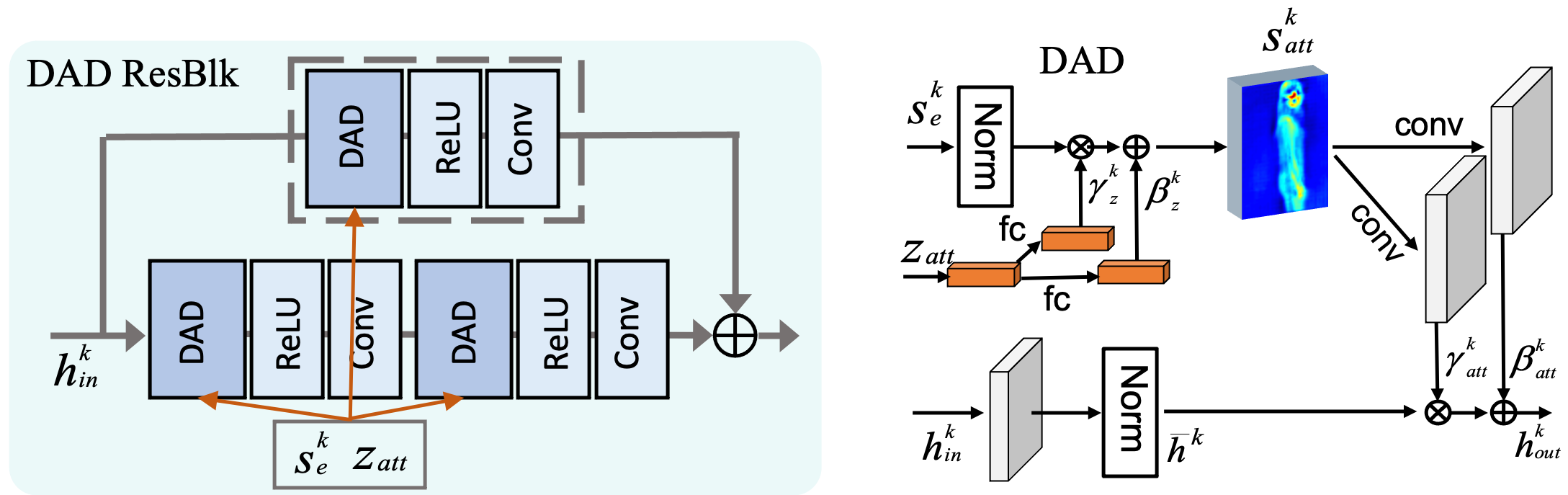}
\caption{Detailed views of (left) a DAD ResBlk; (right) a Dual adaptive denormalization layer.}\label{fig:DADnew}
\end{figure*}


\section{Related Work}

\bfsection{Person image synthesis} Benefiting from the success in image synthesis \cite{huang2018multimodal,karras2019style,hu20213dbodynet,tang2021total}, many works have focused on synthesizing person images. PG2 \cite{ma2017pose}
firstly proposed a two-stage GAN architecture \cite{goodfellow2014generative} to generate person images. Esser \etal \cite{esser2018variational} leveraged a variational autoencoder combined with a conditional U-Net \cite{ronneberger2015u} to model the inherent shape and appearance. Siarohin \etal \cite{siarohin2018deformable} used a U-Net based generator with deformable skip connections to handle the pixel-to-pixel
misalignments between different poses. Zhu \etal \cite{zhu2019progressive} introduced cascaded Pose-Attentional Transfer Blocks to progressively guide the person image synthesis. \cite{song2019unsupervised,pumarola2018unsupervised} utilized a bidirectional strategy for synthesizing person images in an unsupervised manner. SMIS \cite{zhu2020semantically} proposed to generate diverse person images with semantic grouping and injection. SPG \cite{lv2021learning} and PISE \cite{zhang2021pise} deal with the human generation task by predicting semantic parsing masks as guidance. CASD~\cite{zhou2022cross} and 
NTED~\cite{ren2022neural} introduces the attention based style distribution module that learns representative features. ADGAN \cite{men2020controllable} introduced a controllable way to synthesize person images that allow for attribute editing. However, it lacks efficient ways to learn disentangled representation for efficient person editing. Our method overcomes these challenges with a novel encoder architecture and a better training strategy. 


\bfsection{Disentangled representation learning} Generating disentangled representations \cite{xiao2022learning,wang2021one,zhang2021disentangled,Xu_2020_WACV,zhou2014smart,xu2020adaptively} is essential for tasks involving attribute editing. InfoGAN \cite{chen2016infogan} applies information regularization to obtain interpretable latent representations. StyleGAN \cite{karras2019style} is able to synthesize impressive images in high-resolution by integrating adaptive instance normalization layers \cite{huang2017arbitrary} in a new generator architecture that learns disentangled latent representations. DPIG \cite{Ma_2018_CVPR} learns pose and appearance representations separately. Our model is the first attempt to learn disentangled latent semantic representations with transformers for controllable person image generation.

\bfsection{Visual Transformers} Transformer has achieved impressive success in object detection \cite{carion2020end,zhu2020deformable} and semantic segmentation \cite{xie2021segmenting,chen2021transunet,zheng2020rethinking}. Transformer \cite{vaswani2017attention} introduces a new attention mechanism that has been successfully applied in various vision tasks \cite{chen2020pre,he2021transreid,han2020survey}. Similar to non-local neural networks \cite{wang2018non,jung2020hide,vaswani2017attention}, the transformer directly works on sequences of image patches to aggregate information. ViT \cite{dosovitskiy2020image} cuts the image into small patches and globally attends all the patches at every transformer layer. The PVT \cite{wang2021pyramid} introduces a versatile backbone for dense prediction. Our model takes transformer layers to extract disentangled representations for controllable person generation.

\bfsection{Curriculum learning}
Inspired by the human learning process, Bengio \etal \cite{bengio2009curriculum} proposed curriculum learning which starts from a relatively easy task and gradually increases the difficulty of training. It benefits both performance improvement and speed of convergence in various deep learning tasks such as weakly supervised object detection \cite{shi2016weakly}, image captioning \cite{ren2017deep,dong2021dual}, and video application \cite{Xiao:TMM2020}. We exploit curriculum learning by scheduling the difficulty according to our proposed random mask-agnostic strategy.

\section{Proposed Approach}
We propose a novel framework DRL-CPG to synthesize person images with user-controlled human attributes, such as pose, head,
upper clothes, and pants. As illustrated in Figure~\ref{fig:overview}, our DRL-CPG takes a collection of source person image $I_s$ and the corresponding semantic mask $M_s$ to provide component attributes, and a target keypoint-based pose $P_t$ to provide the target pose attribute. The framework consists of two components, \ie, an attribute encoder with transformer trained in a manner of curriculum learning, and an attribute decoder with dual adaptive denormalization for attribute transfer.

\subsection{Attribute Encoder with Transformers}\label{sec:encoder_with_transformers}
To learn disentangled representation for controllable person generation, we propose a novel attribute encoder with transformers to generate the intermediate representation set. 
The attribute encoder flattens region components in source images and supplements them with a positional encoding before
passing it into a transformer encoder. A transformer decoder then takes as input of the encoded feature map, and searches for matching with a small fixed number of learned component embeddings, which we call ``{\em component queries}". The overall attribute encoder architecture consists of a transformer encoder and a transformer decoder that generates the final feature representations. 

\bfsection{Transformer encoder} Our transformer encoder is adopted from PVT \cite{wang2021pyramid}. Given an input image of size $(H,W, 3)$, along with fixed positional encodings to compensate for the missing spatial information, the transformer encoder generates a $(\frac{H}{8},\frac{W}{8}, C)$ feature map, which is then flattened into a sequence of $( \frac{H}{8}\times\frac{W}{8}, C)$. The $H, W, C$ refer to the height, width, and channel dimensions. For simplification, we denote the size of this flattened feature as $(T, C)$, where $T = \frac{H}{8}\times\frac{W}{8}$. The encoder is composed of stacked transformer layers, each of which consists of a multi-head self-attention module and a feed-forward network.

\bfsection{Transformer decoder}
The goal of our transformer decoder is to take input a set of learnable component embedding as ``{\em component queries}", denoted by $E_{c}\in R^{N\times L}$, the encoded feature as key and value, denoted by $E_{f} \in R^{T\times C}$, and output a new component embedding. $L$ stands for the feature dimension and $N$ refers to the number of components. This can be formulated as below:
{\small
\begin{align}
\label{eqx}
\texttt{Qattn}(E_{c})&=\texttt{softmax}((E_{c}W_{\theta})^T(E_{f}W_{\phi})),\\
Z^c_{att} &= (\texttt{Qattn}(E_{c}) (E_{f}W_z)) W_{c},
\end{align}
}
where each $W$ represents the parameters of MLP and $\texttt{Qattn}$ is a query attention map used to highlight different individual components. Finally, the output $Z^c_{att} \in R^{N\times L}$ is taken as the intermediate feature with the dimension of $L$. We take the positioning index so that each item in $Z^c_{att}$ learns representative information for a specific person component. This is different from most previous transformers working at the instance level. We use the transformers to learn localizable features within objects (different parts in the same instance).

\subsection{Random Component Mask-Agnostic}\label{sec:curriculum_learning}
In order to learn the texture and shape representation of each component, existing methods take the semantic mask to separate each component at the pixel level and let the encoder learn the representation of each separated component. This tends to learn entangled representation since the component shape is correlated with other components. For instance, the length of the uncovered arm is directly determined by the upper cloth. As a result, the original cloth item that is not completely removed causes problems in learning the latent representation.

\begin{figure}[t!]
\centering     
\includegraphics[width=1.\linewidth]{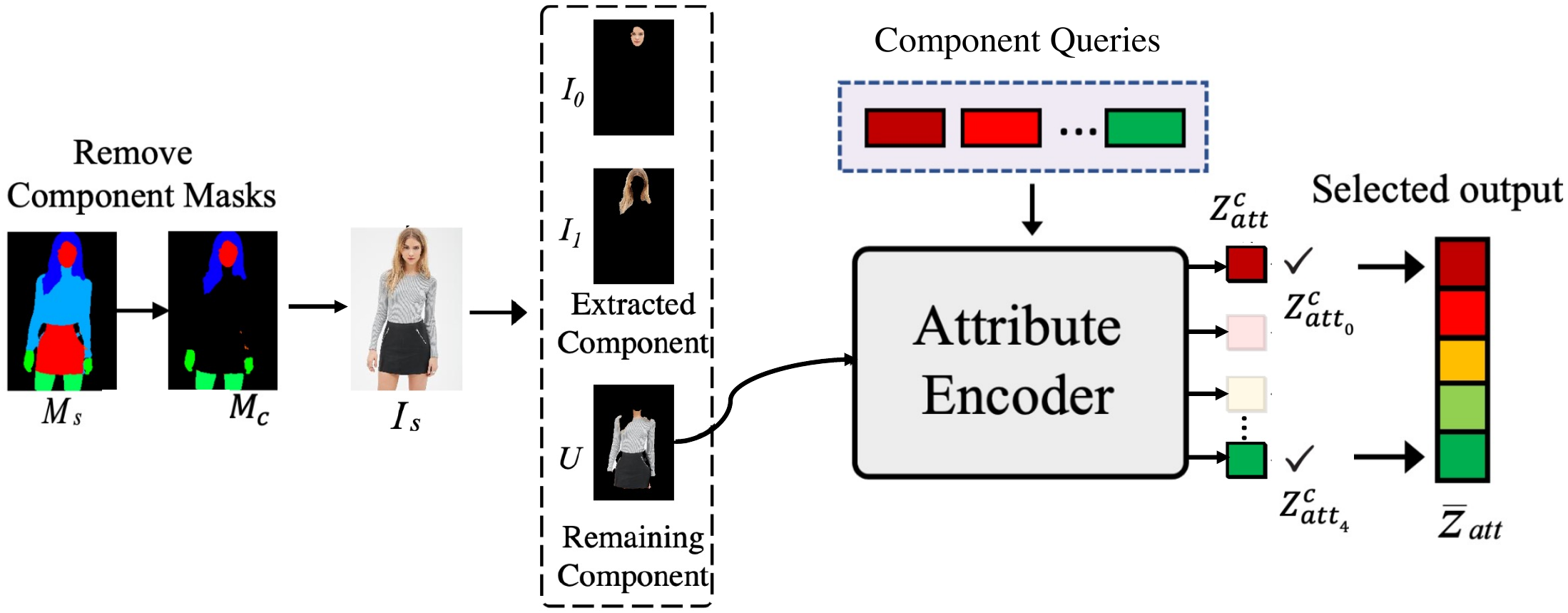}
\caption{Attribute encoder works at random component mask-agnostic strategy.}\label{fig:Mask-agnostic}
\end{figure}

To address this, we propose a random component mask-agnostic strategy to train the model, which truly eliminates the correlation of each component and promotes the model to learn disentangled representation. The workflow is shown in Figure \ref{fig:Mask-agnostic}. Given a semantic mask $M_s$ containing $N$ attribute regions, \eg, head, upper clothes, skirt, and pants, we first randomly generate a set of indexes indicating the need to remove the component masks from $M_s$. After removing the component masks, we extract components based on the remaining component masks $M_c$. This creates two types of components, denoted by $\{ I_c, U\}$, where $I_c= M_c*I_s$ given that $M_c$ is not removed. $U$ is termed as the mask-agnostic component obtained as $U= I - \sum I_c$, where superscript $c$ refers to the $c$-th component. Then we treat these two types of components equally and feed them into the attribute encoder. For each component, the encoder produces $Z^c_{att} \in R^{N\times L} $. We then pick out an item from $Z^c_{att}$ as the representation of each individual component. This can be described as $Z^c_{{att}_j} =S_j(Z^c_{att}) \in R^{1\times L}$, where $S_j(\cdot)$ represents the feature selection that takes the $j$-th row as output. Finally, we concatenate the selected representations to get $\bar{Z}_{att} = \texttt{concat}(Z^c_{{att}_j} ) \in R^{N \times L}$. $\bar{Z}_{att}$ is the final latent representation of the person image $I_s$. 

Our random component mask-agnostic strategy introduces different learning tasks in terms of recognition difficulty. If all the component masks are removed, we feed the complete person image into the attribute encoder. The attribute encoder needs to recognize each component without supervision. This is the most difficult task. If parts of the component masks are removed, the remaining component masks are used in extracting components. If all the component masks are available, each component will be separated at the image level. This is an easy task for the attribute encoder to produce representations for each extracted component. According to different levels of difficulty, we further introduce the curriculum learning strategy to train our model.

\bfsection{Curriculum learning} 
Let the number of available component masks $k$ indicate the difficulty of a task. $k=N$ indicates an easy task where a complete semantic mask is available to separate different components. While $k=0$ indicates a hard task where no semantic mask is available to separate each component. As we can see, the random component mask-agnostic strategy increases the difficulty in learning latent representations and thus enhances the encoder to recognize each component with or without component masks. To further alleviate the training difficulty in early steps, we adopt a curriculum learning scheme\cite{bengio2009curriculum}. At the early training stage, our network is given fully separated components (\ie $k = N$). After $\alpha$ epochs, we start to take in the random component mask-agnostic strategy to randomly remove component masks (\ie $k < N$). 
\begin{figure}[ht!]
\captionsetup[subfigure]{justification=raggedright, singlelinecheck=false, labelformat=empty} 
\centering
  \includegraphics[height=0.22\textwidth]{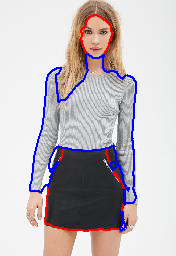}
  \hspace{-2.5pt} 
  \includegraphics[height=0.22\textwidth]{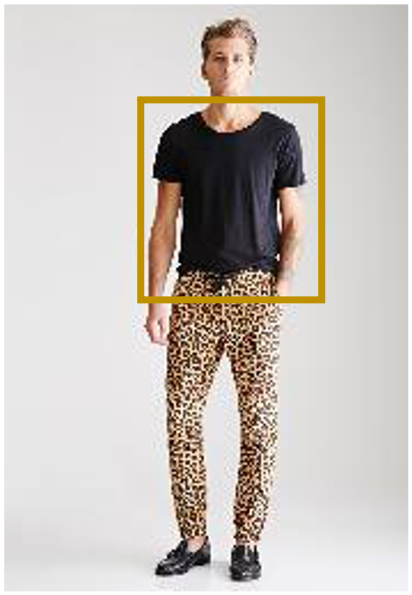} 
\includegraphics[height=0.22\textwidth]{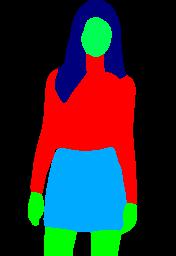} 
  \hspace{-4pt}
 \includegraphics[height=0.22\textwidth]{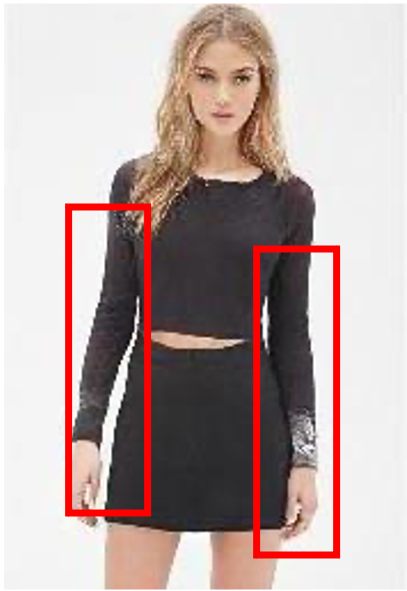} 
    \includegraphics[height=0.22\textwidth]{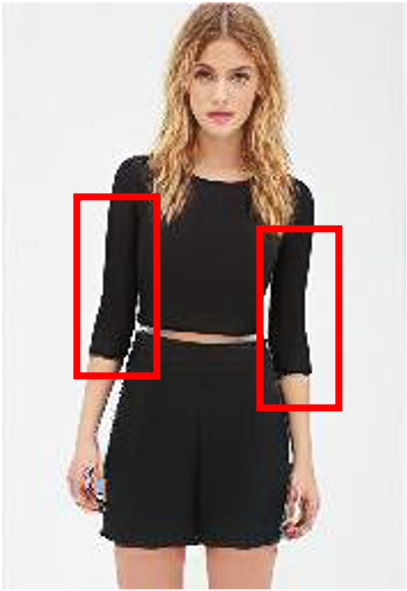} 
  \hspace{-2.5pt}  
  \includegraphics[height=0.22\textwidth]{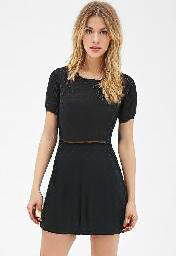}\\ 
    \vspace{1pt}
  \includegraphics[height=0.22\textwidth]{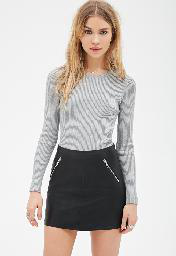}
  \hspace{-2.5pt}
  \includegraphics[height=0.22\textwidth]{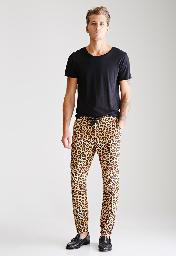}
  \includegraphics[height=0.22\textwidth]{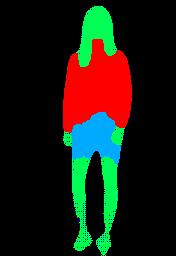} 
  \hspace{-4pt}
  \includegraphics[height=0.22\textwidth]{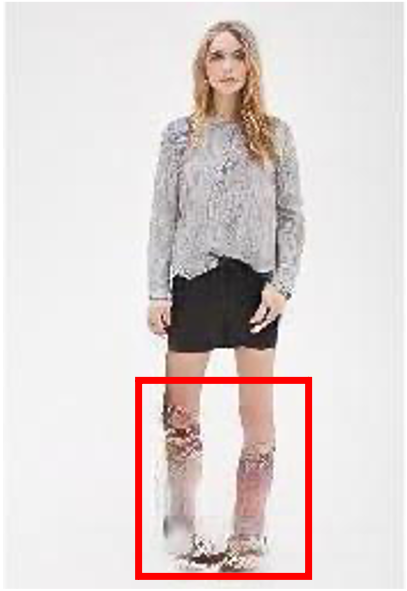} 
  \includegraphics[height=0.22\textwidth]{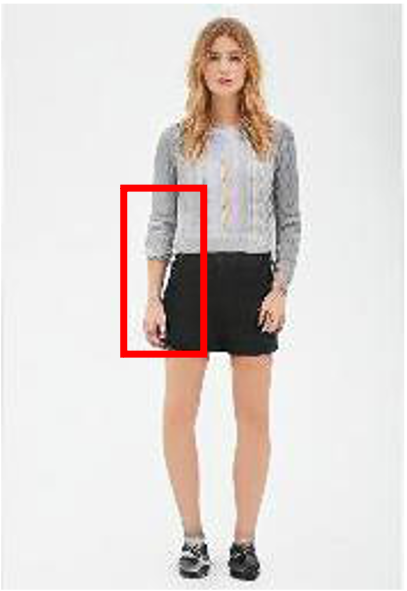}
  \hspace{-2.5pt}  
  \includegraphics[height=0.22\textwidth]{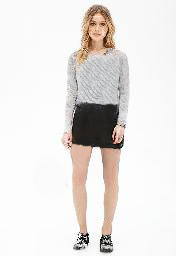} 
  \begin{subfigure}{\linewidth}
  \caption[l]{  \hspace{ 12pt} Base \hspace{ 16pt}  Cloth \hspace{ 6pt} 
  PISE  {\scriptsize + Predicted mask}
  \hspace{ 4pt} ADGAN \hspace{ 9pt} Ours} 
  \end{subfigure}
  \vspace{-0.5cm}		
   \caption{\label{fig:trans_intro} Comparison between different methods. (Top) The results of component attribute transfer; (Bottom) The results of pose transfer.}
\end{figure}
 
 
{\noindent\bf Performance preview}. To demonstrate how well our approach works, we compare three performances on person generation task in Figure \ref{fig:trans_intro}. In this case, we try to transfer the upper cloth from the source person to the base. We observe that our DRL-CPG yields better person images with a short sleeve that is consistent with the source cloth, while ADGAN fails to create a consistent sleeve style and PISE creates a long sleeve style following the mask estimated by itself. The predicted problematic mask will introduce holes and artifacts into the generated person. Our model is robust to recognize the underlying boundaries between different components and is able to transfer both the shape and texture of components.

\subsection{Attribute Decoder}
The attribute decoder is to generate person images conditioned on semantic attributes and the target pose. This network consists of a pose-guided structure completion network and a dual adaptive denormalization generator. In the beginning, we take a multilayer perceptron (MLP) module 
to recombine the disentangled features into a more global representation $Z_{att}$.

\bfsection{Pose-guided structure completion network} The previous models generate person images conditioned on the pose, which is defined by several landmarks indicating the positions of each joint. However, this lacks details on the structure, making the person image synthesis an ill-posed problem. In order to create the structural
details, we propose to generate multi-level feature maps based on the input pose. Specifically, we feed the target pose $P_t$ into a UNet-like structure and select the intermediate feature maps $S_{e}^k$ from the $k_{th}$ level feature map of the UNet decoder. These features reflect the spatial structure indicated by the target pose. They are further utilized to guide the person image generation.

\bfsection{Dually adaptive denormalization generator}
Our generator integrates two types of representation, $Z_{att}$ and $S_{e}$, to generate a person image $I_g$. As discussed, the $Z_{att}$ is a one-dimensional semantic representation, while $S_{e}$ is a spatial feature. We propose a novel Dually Adaptive Denormalization (DAD) layer, as shown in Figure~\ref{fig:DADnew}, to integrate them in a more adaptive fashion. Let $h^k_{in} \in R^{C^k\times H^k\times W^k}$ denote one activation map that is fed into the $k_{th}$ layer, with $C^k$ being the number of channels and $H^k\times W^k$ being the spatial dimensions. The output is calculated by
{\small
\begin{align}
\label{eqx}
\begin{split}
h^k_{out}=\gamma^k_{att} \otimes \frac{h^k_{in}-\mu^k}{\sigma^k}+\beta^k_{att},
\end{split}
\end{align}
}
where $\mu^k \in R^{C^k}$ and $\sigma^k \in R^{C^k}$ are the means and standard deviations of the channel-wise activations within $h^k_{in}$; and $\gamma^k_{att}$ and $\beta^k_{att}$ are two modulation parameters used to inject semantic information into the normalized activations. Typically, these two parameters are calculated from two-dimensional semantic segmentation maps. Since our spatial features $S_{e}^k$ contain only structural information, we adopt another modulation operator to adaptively adjust its effective regions, and inject the semantic information learned in $Z_{att}$ into corresponding regions. This produces a semantic feature map $S_{att}^k$. Formally, this can be described as
{\small
\begin{align}
\label{eqx}
\begin{split}
S_{att}^k=\gamma^k_{z} \otimes \frac{S_{e}^k-\mu_e^k}{\sigma_e^k} +\beta^k_{z},
\end{split}
\end{align}
}
where $\mu_e^k$ and $\sigma_e^k$ are the means and standard deviations of the channel-wise activations within $S_{e}^k$; and $\gamma^k_{z}$ and $\beta^k_{z}$ are two modulation parameters both mapped from $Z_{att}$ with two full connection layers. Two convolutional layers are used to generate $\gamma^k_{att}$ and $\beta^k_{att}$ based on $S^k_{att}$.

The DAD Residual Block (DAD ResBlk) is designed as a combination of ``DAD+Relu+Conv" with a residual connection~\cite{he2016deep}. With the attribute representation $Z_{att}$ and the structure feature $S_e$, we cascade DAD residual
blocks to generate the target person $I_g$.

\subsection{Loss Functions} 
The joint loss function is formulated with an adversarial loss $\mathcal{L}_{adv}$, a reconstruction loss $\mathcal{L}_{rec}$,  a perceptual loss $\mathcal{L}_{per}$~\cite{johnson2016perceptual}, and a contextual loss $\mathcal{L}_{ctx}$~\cite{mechrez2018contextual,men2020controllable} as 
{\small
\begin{align}
\label{eqx}
\begin{split}
\mathcal{L} =\mathcal{L}_{adv}+\lambda_{rec}\mathcal{L}_{rec}+\lambda_{per}\mathcal{L}_{per}&+\lambda_{ctx}\mathcal{L}_{ctx}
\end{split}
\end{align}
}
where we set $\lambda_{rec}=1$, $\lambda_{per}=5$ and $\lambda_{ctx}=1$ in our experiments.

\bfsection{Adversarial loss}
We employ an adversarial loss $\mathcal{L}_{adv}$ with discriminators $D_p$ and $D_t$ to help the generator $G$ synthesize the target person image with visual textures similar to the reference one, as well as following the target pose. It penalizes for the distance between the distribution of real pair$(P_t, I_t)$ and the distribution of fake pair $(P_t, I_g)$ containing generated images
{\small
\begin{align}
\label{eqx}
\begin{split}
\mathcal{L}_{adv}=\mathbb{E}[\log( D_t(I_s,I_t) D_p(P_t,I_t))]+  \mathbb{E}[\log ((1-D_t(I_s,I_g))\\(1- D_p(P_t,I_g)))]
\end{split}
\end{align}
}
\bfsection{Reconstruction loss}
We define a reconstruction loss as the pixel-level $L_1$
distance between the target image $I_t$ and the generated image
{\small
\begin{align}
\label{eqx}
\begin{split}
\mathcal{L}_{rec} =||G(I_s,P_t)-I_t||_1 
\end{split}.
\end{align}
}

\begin{figure}[ht!]
\vspace{-0.2cm}
\captionsetup[subfigure]{justification=raggedright, singlelinecheck=false, labelformat=empty}    
 \centering
 
\includegraphics[width=0.8\textwidth]{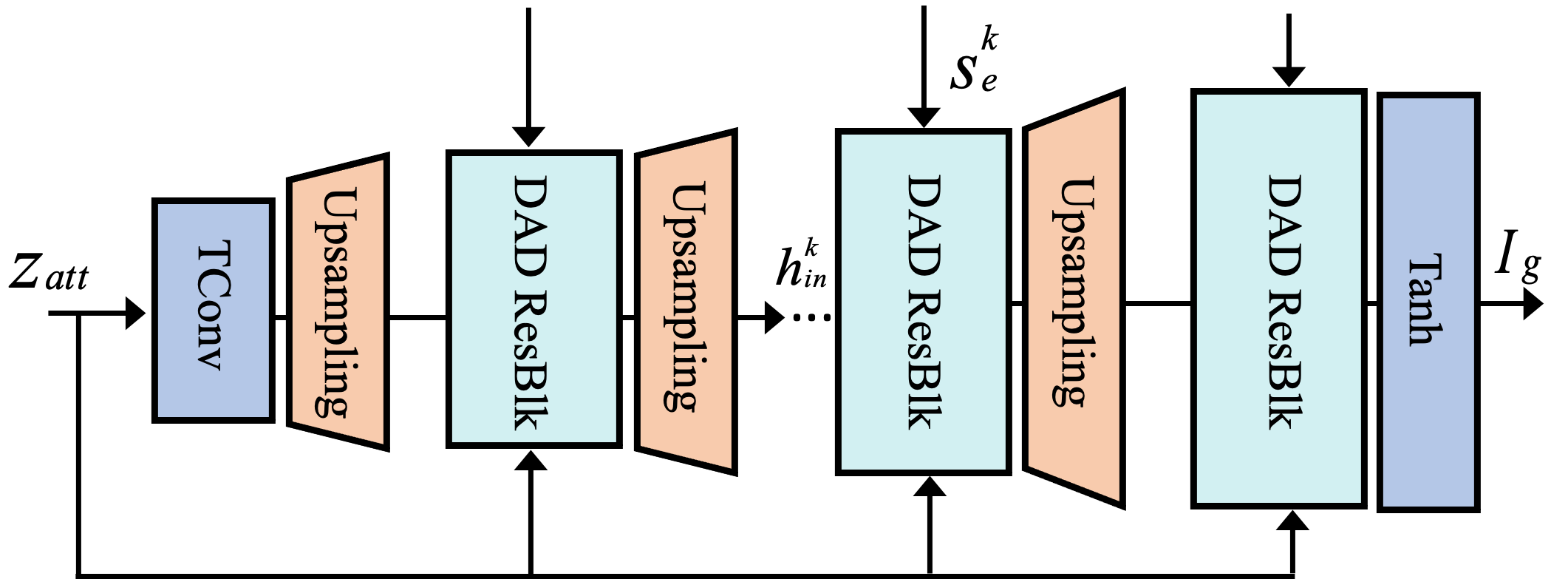}

\caption{Architecture of our DAD generator. TConv refers to the TransposeConvolutional layer.}
\label{fig:decoder}
\vspace{-0.2cm}
\end{figure}

\section{Experiments}\label{sec:exp}
To evaluate the effectiveness of our DRL-CPG, we compete it with two types of person generation methods including (a) pose transfer method \ie PG$^2$ \cite{ma2017pose}, DPIG \cite{Ma_2018_CVPR}, Def-GAN \cite{siarohin2018deformable},
PATN \cite{zhu2019progressive} and SPG \cite{lv2021learning}; (b) both pose transfer and component attribute transfer method \ie ADGAN \cite{men2020controllable} and PISE \cite{zhang2021pise}. Following the data pre-processing manner in \cite{zhu2019progressive,men2020controllable}, from the DeepFashion \cite{liu2016deepfashion} dataset we take 101,966 pairs of images for training and 8,750 pairs for testing. We also evaluate the performance on the Market1501 dataset~\cite{zheng2015scalable} which contains 12,936 training images and 19,732 testing images.
For each image, we acquire the semantic map of a person image with the human parser \cite{gong2017look}. 
All the images are with a resolution of $256\times 256$. 
\begin{figure*}[ht!]
\captionsetup[subfigure]{justification=raggedright, singlelinecheck=false, labelformat=empty} 
\centering
 \includegraphics[width=0.066\textwidth]{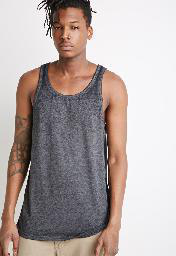}
 \hspace{-5pt}
\includegraphics[width=0.066\textwidth]{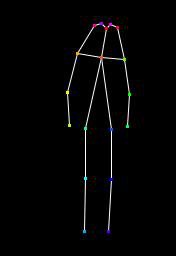}
\hspace{-5pt}
\includegraphics[width=0.066\textwidth]{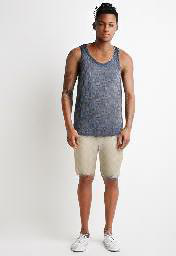}
\hspace{-5pt}
\includegraphics[width=0.066\textwidth]{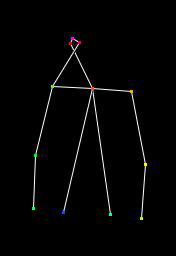}
\hspace{-5pt}
\includegraphics[width=0.066\textwidth]{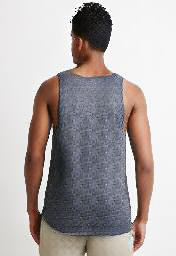}
\hspace{-5pt}
\includegraphics[width=0.066\textwidth]{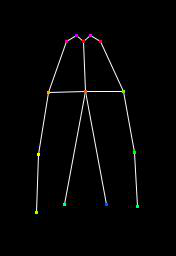}
\hspace{-5pt}
\includegraphics[width=0.066\textwidth]{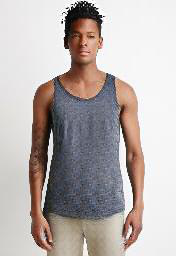}
\hspace{-4pt}
 \includegraphics[width=0.066\textwidth]{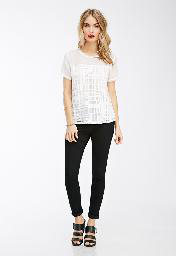}
 \hspace{-5pt}
\includegraphics[width=0.066\textwidth]{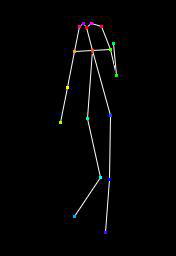}
\hspace{-5pt}
\includegraphics[width=0.066\textwidth]{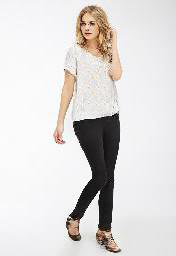}
\hspace{-5pt}
\includegraphics[width=0.066\textwidth]{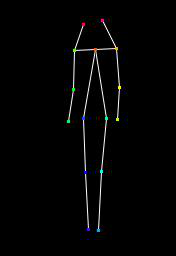}
\hspace{-5pt}
\includegraphics[width=0.066\textwidth]{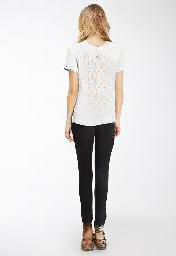}
\hspace{-5pt}
\includegraphics[width=0.066\textwidth]{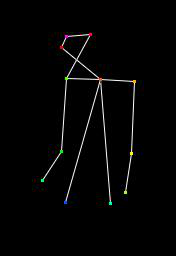}
\hspace{-5pt}
\includegraphics[width=0.066\textwidth]{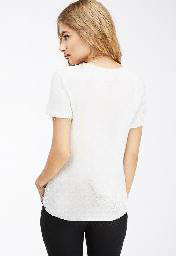}

\raisebox{0.00001\textwidth }{\makebox[0.066\textwidth]{\parbox{0.0001\linewidth}{}}}
\hspace{-5pt}  
\includegraphics[width=0.066\textwidth]{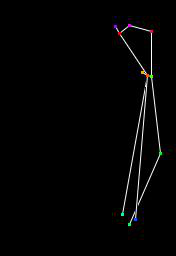}
\hspace{-5pt}
\includegraphics[width=0.066\textwidth]{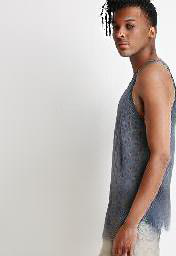}
\hspace{-5pt}
\includegraphics[width=0.066\textwidth]{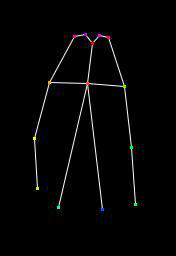}
\hspace{-5pt}
\includegraphics[width=0.066\textwidth]{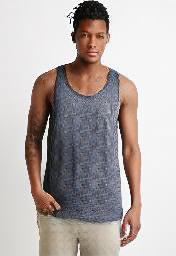}
\hspace{-5pt}
\includegraphics[width=0.066\textwidth]{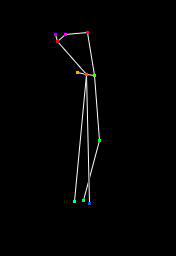}
\hspace{-5pt}
\includegraphics[width=0.066\textwidth]{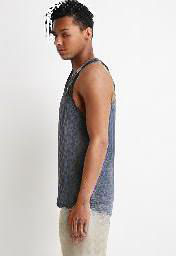}
\hspace{-4pt}
\raisebox{0.00001\textwidth }{\makebox[0.066\textwidth]{\parbox{0.0001\linewidth}{}}}
\hspace{-5pt}
\includegraphics[width=0.066\textwidth]{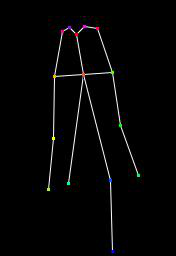}
\hspace{-5pt}
\includegraphics[width=0.066\textwidth]{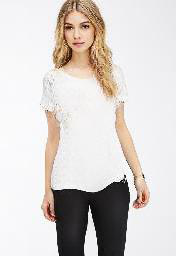}
\hspace{-5pt}
\includegraphics[width=0.066\textwidth]{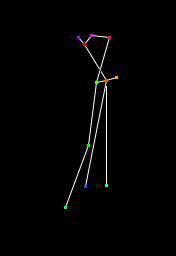}
\hspace{-5pt}
\includegraphics[width=0.066\textwidth]{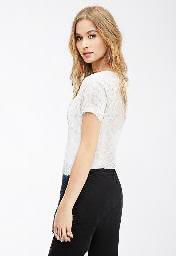}
\hspace{-5pt}
\includegraphics[width=0.066\textwidth]{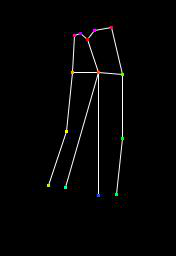}
\hspace{-5pt}
\includegraphics[width=0.066\textwidth]{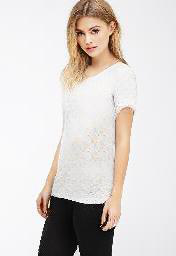}
  \begin{subfigure}{\linewidth}
  \caption[l]{ \hspace{ 2pt}  Person \hspace{ 16pt}  Pose \hspace{ 12pt}  Ours  \hspace{ 12pt}  Pose \hspace{ 12pt}  Ours \hspace{ 12pt}  Pose \hspace{ 12pt}  Ours  \hspace{ 12pt}  Person\hspace{ 12pt}  Pose \hspace{ 12pt}  Ours\hspace{ 12pt}  Pose \hspace{ 12pt}  Ours\hspace{ 16pt}  Pose \hspace{ 16pt}  Ours} 
  \end{subfigure}
\caption{Results of synthesizing person images in arbitrary poses. }\label{fig:arbitrarypose}
\vspace{-0.3cm}
\end{figure*}

\begin{figure*}[ht!]
\captionsetup[subfigure]{justification=raggedright, singlelinecheck=false, labelformat=empty} 
\centering
\begin{subfigure}[t]{1.\textwidth}
  \centering
\includegraphics[height=0.128\textwidth]{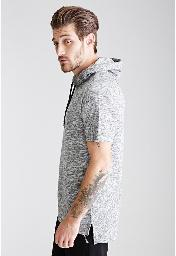}
  \hspace{-5pt}
\includegraphics[height=0.128\textwidth]{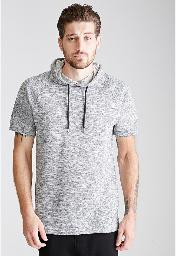}
  \hspace{-5pt}
\includegraphics[height=0.128\textwidth]{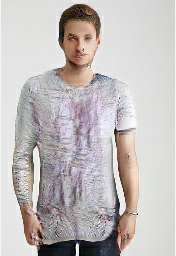}
  \hspace{-5pt}
\includegraphics[height=0.128\textwidth]{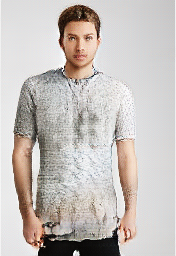}
  \hspace{-5pt}
\includegraphics[height=0.128\textwidth]{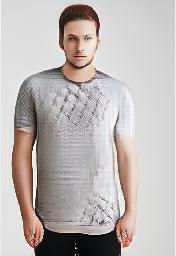}
  \hspace{-5pt}
\includegraphics[height=0.128\textwidth]{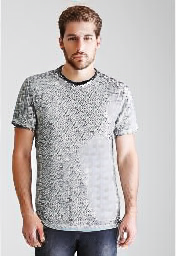}
  \hspace{-5pt}
  \includegraphics[height=0.128\textwidth]{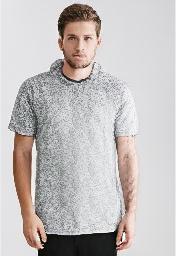}
  \hspace{-5pt} 
  \includegraphics[height=0.128\textwidth]{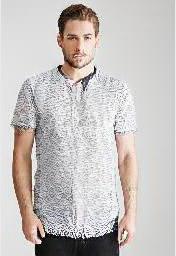} 
  \hspace{-5pt}
  \includegraphics[height=0.128\textwidth]{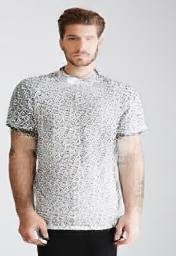} 
  \hspace{-5pt}
  \includegraphics[height=0.128\textwidth]{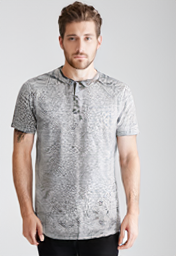}   
  \hspace{-5pt} 
\includegraphics[height=0.128\textwidth]{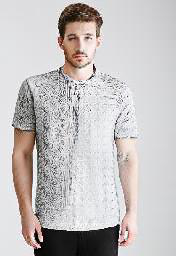}
\end{subfigure} \\
\vspace{3pt}
\begin{subfigure}[t]{1.\textwidth}
 \centering    
\includegraphics[height=0.128\textwidth]{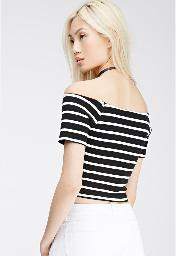}
  \hspace{-5pt}
\includegraphics[height=0.128\textwidth]{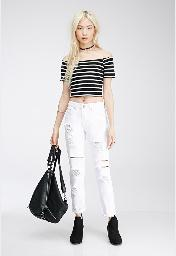}
  \hspace{-5pt}
\includegraphics[height=0.128\textwidth]{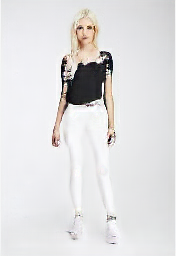}
  \hspace{-5pt}
\includegraphics[height=0.128\textwidth]{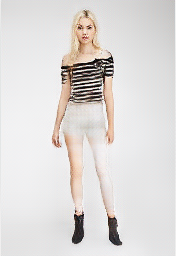}
  \hspace{-5pt}
\includegraphics[height=0.128\textwidth]{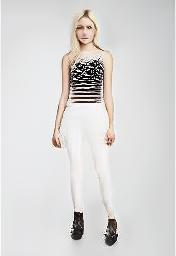}
  \hspace{-5pt}
\includegraphics[height=0.128\textwidth]{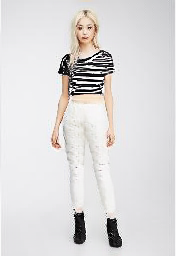}
  \hspace{-5pt}
 \includegraphics[height=0.128\textwidth]{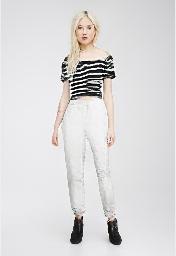}
  \hspace{-5pt}
 \includegraphics[height=0.128\textwidth]{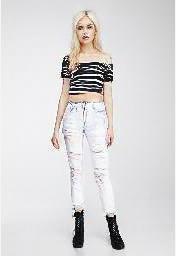} 
  \hspace{-5pt}
  \includegraphics[height=0.128\textwidth]{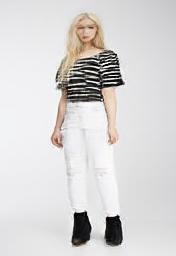} 
  \hspace{-5pt}
  \includegraphics[height=0.128\textwidth]{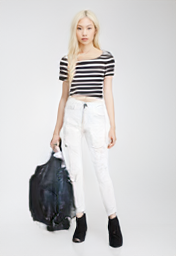}   
  \hspace{-5pt} 
\includegraphics[height=0.128\textwidth]{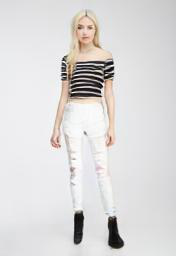}
   \end{subfigure}   \\
    \vspace{3pt}
  \begin{subfigure}[t]{1.\textwidth}
  \centering  
  \includegraphics[height=0.128\textwidth]{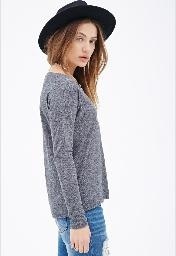}
  \hspace{-5pt} 
  \includegraphics[height=0.128\textwidth]{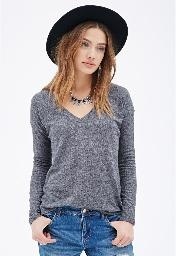}
  \hspace{-5pt}
  \includegraphics[height=0.128\textwidth]{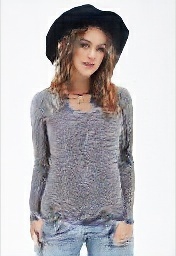}
  \hspace{-5pt}
  \includegraphics[height=0.128\textwidth]{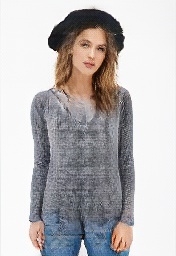}
  \hspace{-5pt}
  \includegraphics[height=0.128\textwidth]{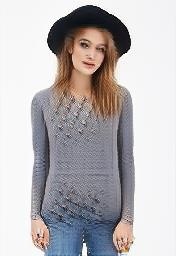}
  \hspace{-5pt}
  \includegraphics[height=0.128\textwidth]{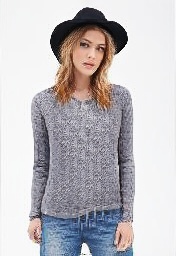}
  \hspace{-5pt}
  \includegraphics[height=0.128\textwidth]{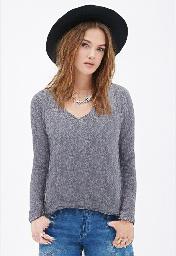}
  \hspace{-5pt}
  \includegraphics[height=0.128\textwidth]{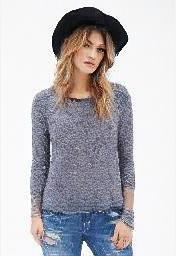} 
  \hspace{-5pt}
  \includegraphics[height=0.128\textwidth]{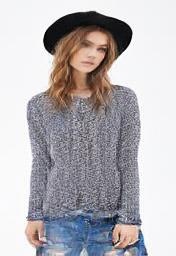} 
  \hspace{-5pt}
  \includegraphics[height=0.128\textwidth]{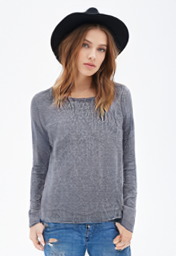}   
  \hspace{-5pt} 
  \includegraphics[height=0.128\textwidth]{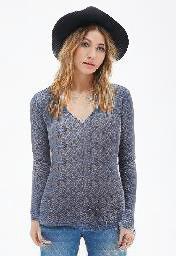}
      
\caption[l]{  \hspace{ 8pt} Base \hspace{ 30pt} GT \hspace{ 18pt} DPIG \hspace{ 14pt} Def-GAN \hspace{ 14pt} PATN \hspace{ 16pt} ADGAN \hspace{ 16pt} SPG \hspace{ 18pt} PISE \hspace{ 22pt}   CASD \hspace{ 18pt}  NTED \hspace{ 22pt} Ours}
\end{subfigure}
\caption{Qualitative comparison with baseline methods.}
\label{fig:pose}
\vspace{-0.3cm}
\end{figure*}

\begin{figure}[ht!]
\includegraphics[width=0.16\textwidth]{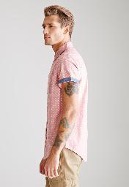}
\hspace{-4.5pt}
\includegraphics[width=0.16\textwidth]{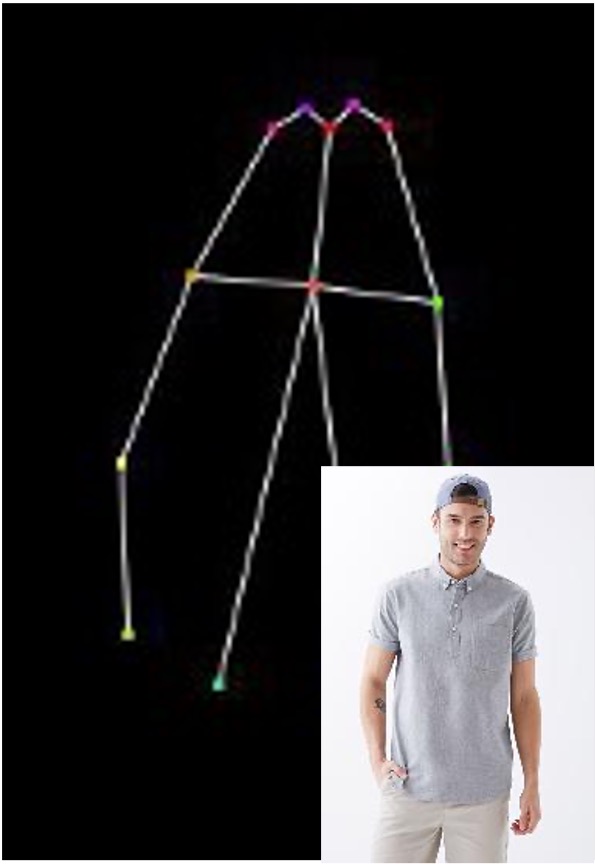}
\hspace{-4.5pt}
\includegraphics[width=0.16\textwidth]{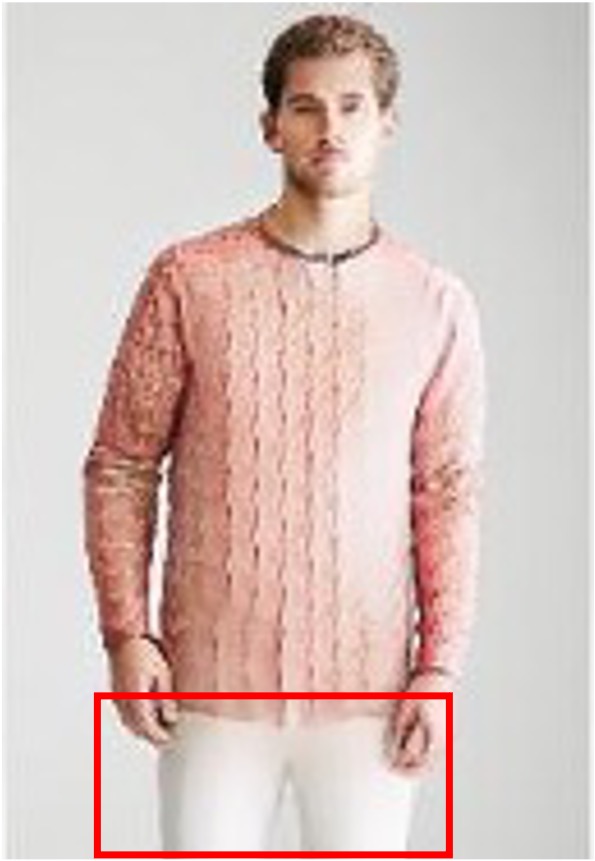}
\hspace{-4.5pt}
\includegraphics[width=0.16\textwidth]{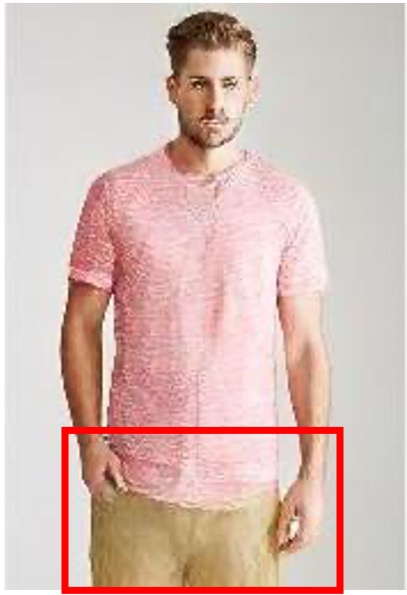}
\hspace{-4.5pt}
\includegraphics[width=0.16\textwidth]{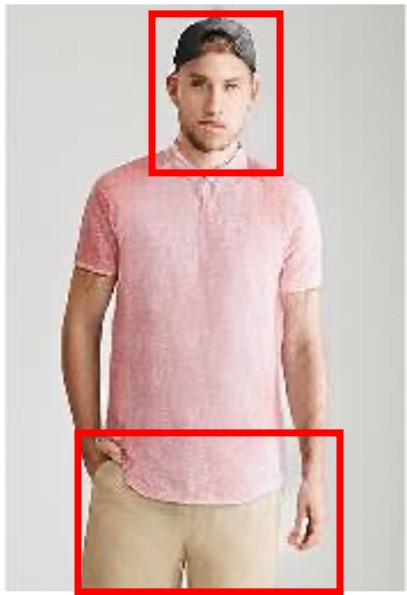}
\hspace{-4.5pt}
\includegraphics[width=0.16\textwidth]{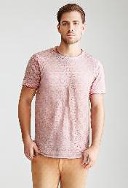}

\includegraphics[width=0.16\textwidth]{rebuttal_pose/1}
\hspace{-4.5pt}
\includegraphics[width=0.16\textwidth]{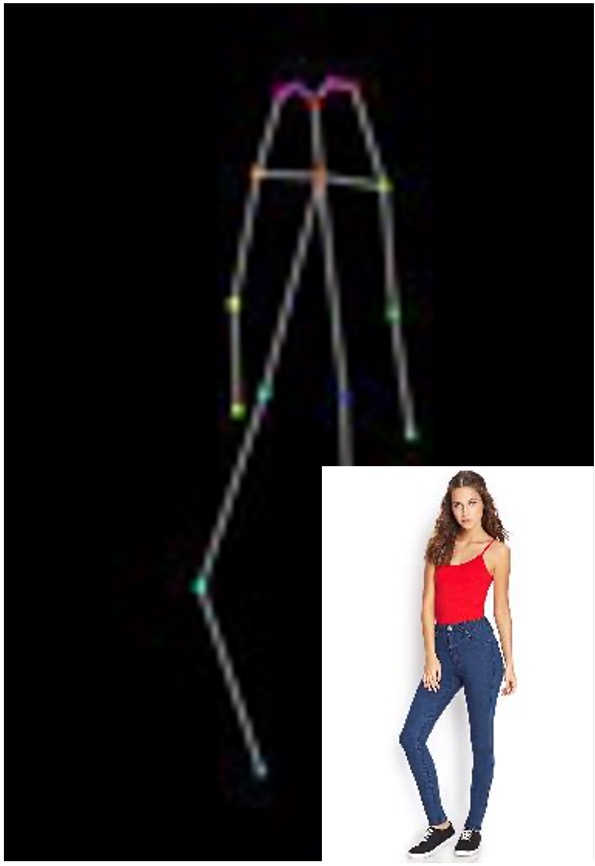}
\hspace{-4.5pt}
\includegraphics[width=0.16\textwidth]{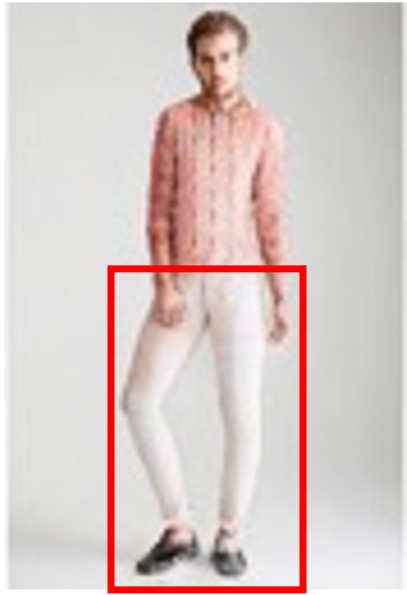}
\hspace{-4.5pt}
\includegraphics[width=0.16\textwidth]{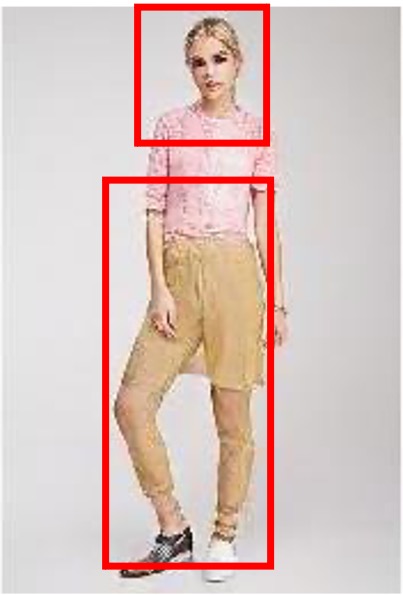}
\hspace{-4.5pt}
\includegraphics[width=0.16\textwidth]{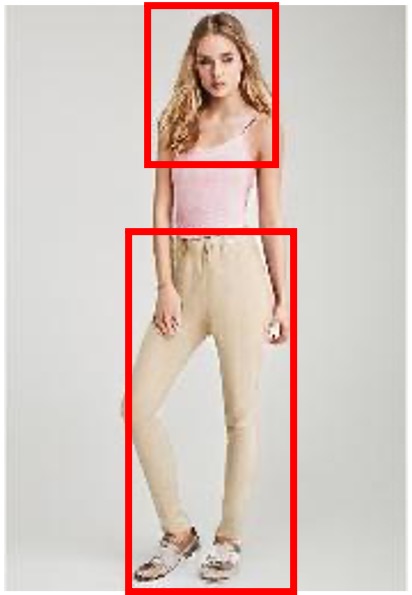}
\hspace{-4.5pt}
\includegraphics[width=0.16\textwidth]{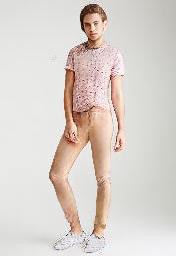}

{ \hspace{ 0pt}  Person \hspace{ 12pt} Pose  \hspace{ 6pt} ADGAN\hspace{ 10pt} PISE\hspace{ 18pt} SPG\hspace{ 15pt} Ours}
\caption{Swap pose transfer results.}
\label{fig:swap_pose}  
\vspace{-0.3cm}
\end{figure}

\begin{table}[ht!]
\centering
\caption{Quantitative comparison on the DeepFashion dataset. CAT is component attribute transfer.}
\begin{tabular}{l|c|c|c|c}
\hlineB{2} 
 Method&CAT& FID$\downarrow$  & SSIM$\uparrow$ &LPIPS $\downarrow$ \\
\hline
PG$^2$~\cite{ma2017pose}&N&23.202&0.773&0.259   \\  
DPIG~\cite{Ma_2018_CVPR}&N& 21.323 &0.745&0.246 \\  
Def-GAN~\cite{siarohin2018deformable}&N&18.475 &0.760&0.2330 \\  
PATN~\cite{zhu2019progressive}&N&20.739& 0.773&0.2533   \\
SPG~\cite{lv2021learning}&N&12.243&0.790 &0.2105 \\
\hline
ADGAN~\cite{men2020controllable}&Y& 14.460&0.772&0.2256  \\
PISE~\cite{zhang2021pise}&Y&13.610&0.778& 0.2059 \\
CASD~\cite{zhou2022cross}&Y& 13.939& 0.768&  0.2174 \\
NTED~\cite{ren2022neural}&Y& {\color{red}9.216}& {\color{blue}0.781} & {\color{red}0.1961}\\
\hline
DRL-CPG (Ours)&Y&{\color{blue} 13.514}&{\color{red} 0.792}&{\color{blue} 0.2027}\\
\hlineB{2} 
\end{tabular}
\label{tab:pose}
\vspace{-0.3cm}
\end{table}

\bfsection{Configuration of Our DRL-CPG Networks}
In the transformer encoder, the embedding dim,  number of layers, and MLP ratio are set to be 256, 4, and 2, respectively. The dimension of component attribute embedding $L$ is set to $64$. For each human image, we separate it into $N = 8$ different semantic components, \ie hair, head, arm, upper cloth, leg, pant, background, and skirt. A basic Conv layer contains a $3\times 3$ convolution operation, a normalization, and a ReLU activation sequentially. Our pose-guided structure completion network consists of 7 Conv layers with stride size 2 and 7 Deconv layers with stride size 2. Our dually adaptive denormalization generator consists of 7 up-sampling layers with stride size 2. Each of them is followed by one DAD ResBlk. The structure of our DAD generator is shown in Figure \ref{fig:decoder}.

\bfsection{Implementation Details}
We adopt Adam optimizer \cite{kingma2014adam} to train our model for 100 epochs. The initial learning rate is set to 0.0001 and linearly decayed to 0 after 60 epochs. Without using a curriculum learning strategy (such as Base + MA model), we start to randomly remove component masks (\ie $k < N$) from the first epoch. Under curriculum learning strategy (such as Base + MA + CL model and our DRL-CPG model), our network is given fully separated components (\ie $k = N$) at the early training stage. We then randomly remove component masks (\ie $k < N$) after $\alpha$ epochs. In our experiment, we take $\alpha=50$. At the testing stage, our DRL-CPG model takes use of all segmentation masks to generate the best person images. We also observe that our trained Base + MA + CL model can generate decent person images and handle component attribute transfer without using segmentation masks to extract components. This proves the effectiveness of our proposed random component mask-agnostic strategy in learning disentangled representation. 






\bfsection{Evaluation metrics}
We conduct several metrics to evaluate the performance of the human pose transfer task. With the source and target image pairs available, we calculate the metrics scores, including Learned Structural Similarity Index Measure (SSIM \cite{wang2004image}), Fr\'{e}chet Inception Distance (FID \cite{heusel2017gans}) and Learned Perceptual Image Patch Similarity (LPIPS \cite{zhang2018perceptual}), to compute the distance between the generated images and the corresponding ground-truth images.

\subsection{Pose Transfer}
Given a source person image and target pose extracted from another person image, 
the task of pose transfer is to generate a natural and realistic person in the shape of the target pose while preserving the person's identity.

\begin{table}[t!]
\centering
\caption{Quantitative comparison on the Market1501 dataset. CAT indicates component attribute transfer.} 
\begin{tabular}{l|c|c|c|c} \hlineB{2}
 Method   & CAT & FID $\downarrow$ & SSIM $\uparrow$ & LPIPS $\downarrow$ 
 \\ 
 \hline
 SPG~\cite{lv2021learning}  & N & {\color{blue}23.331} & {\color{red}0.315} & {\color{red}0.2779} \\
 \hline
 ADGAN~\cite{men2020controllable}  & Y & 26.784 & 0.261 & 0.3162 \\
 PISE~\cite{zhang2021pise}  & Y & 24.852 & 0.273 & 0.3073 \\
 \hline
 DRL-CPG (Ours) &   Y & {\color{red}22.517} & {\color{blue}0.310} & {\color{blue}0.2789} \\
\hlineB{2}
\end{tabular}~\label{tab:exp_Market1501}
\vspace{-0.1cm}
\end{table}

\begin{figure}[t!]
 \vspace{-0.0cm}
\captionsetup[subfigure]{justification=raggedright, singlelinecheck=false, labelformat=empty}    
 \centering
\begin{subfigure}{\textwidth}
\vspace{-0.3cm}
\centering
\includegraphics[width=0.145\textwidth]{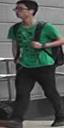}
\hspace{-4pt}
\includegraphics[width=0.145\textwidth]{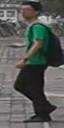}
\hspace{-4pt}
\includegraphics[width=0.145\textwidth]{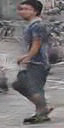}
\hspace{-4pt}
\includegraphics[width=0.145\textwidth]{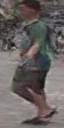}
\hspace{-4pt}
\includegraphics[width=0.145\textwidth]{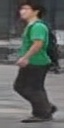}
\hspace{-4pt}
\includegraphics[width=0.145\textwidth]{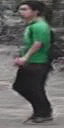}

\includegraphics[width=0.145\textwidth]{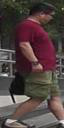}
\hspace{-4pt}
\includegraphics[width=0.145\textwidth]{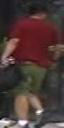}
\hspace{-4pt}
\includegraphics[width=0.145\textwidth]{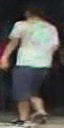}
\hspace{-4pt}
\includegraphics[width=0.145\textwidth]{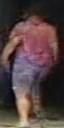}
\hspace{-4pt}
\includegraphics[width=0.145\textwidth]{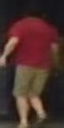}
\hspace{-4pt}
\includegraphics[width=0.145\textwidth]{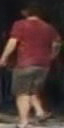}

  \caption[l]{  \hspace{ 10pt} Source \hspace{ 10pt} Target \hspace{ 8pt} ADGAN\hspace{ 10pt} PISE \hspace{10pt} SPG \hspace{ 18pt} Ours}
   \vspace{-0.05cm}
   
\end{subfigure}
\caption{Visual performance comparison on the market-1501 dataset.}
\label{fig:exp_Market1501}
 \vspace{-0.20cm}
\end{figure}

\subsubsection{DeepFashion Dataset} 
We first conduct experiments on the DeepFashion Dataset. 
In Figure \ref{fig:arbitrarypose}, we show some results synthesized by our method. Based on the person image and the target poses, our model generates realistic images.
In Figure \ref{fig:pose}, we compare the performance on the pose transfer task. As we can see, our DRL-CPG produces realistic person images with a better cloth texture and local structure. To further demonstrate the outperformance of our approach, in Figure \ref{fig:swap_pose} we visualize some comparison results where the target poses are from different persons. Our DRL-CPG produces person images maintaining the attributes consistent with the source person, while PISE and SPG wrongly generate female heads when the target poses are adopted from women. Because these two methods only inject the texture code into the estimated masks that determine the final shape of the created person, one of the issues is that the estimated masks are heavily correlated with the poses and thus the generated person preserves features from the target person, such as traces of objects and shapes of components.

In Table \ref{tab:pose}, we list the quantitative results. 
Our DRL-CPG generates the best person images in terms of SSIM score and preserves the similarity of the texture as proved by the LPIPS values; DRL-CPG achieves comparable FID scores, which indicates our method is able to preserve the shape and texture. 

\begin{figure*}[t!]
\captionsetup[subfigure]{justification=raggedright, singlelinecheck=false, labelformat=empty}    
 \centering


\includegraphics[width=0.108\textwidth]{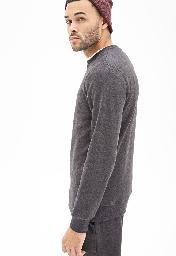}
\hspace{-4.5pt}
\includegraphics[width=0.108\textwidth]{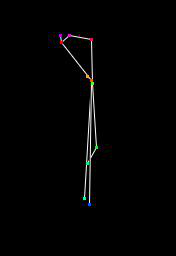}
\hspace{-4.5pt}
\includegraphics[width=0.108\textwidth]{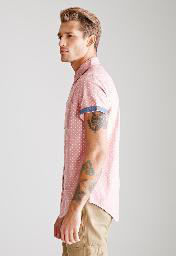}
\hspace{-4.5pt}
\includegraphics[width=0.108\textwidth]{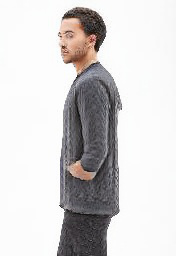}
\hspace{-4.5pt}
\includegraphics[width=0.108\textwidth]{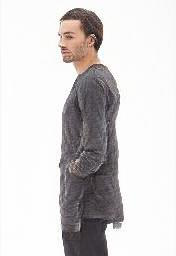}
\hspace{-4.5pt}
\includegraphics[width=0.108\textwidth]{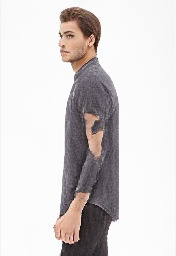}
\hspace{-4.5pt}
\includegraphics[width=0.108\textwidth]{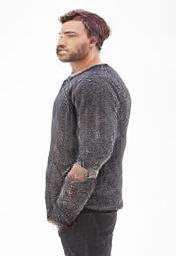}
\hspace{-4.5pt}
\includegraphics[width=0.108\textwidth]{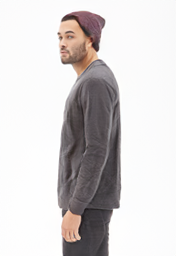}
\hspace{-4.5pt}
\includegraphics[width=0.108\textwidth]{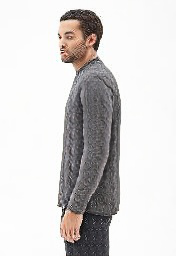}

\includegraphics[width=0.108\textwidth]{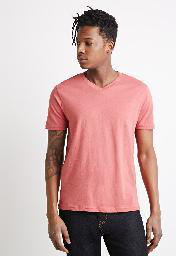}
\hspace{-4.5pt}
\includegraphics[width=0.108\textwidth]{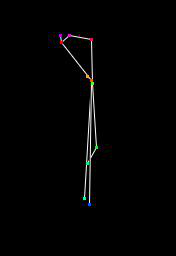}
\hspace{-4.5pt}
\includegraphics[width=0.108\textwidth]{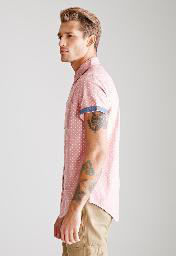}
\hspace{-4.5pt}
\includegraphics[width=0.108\textwidth]{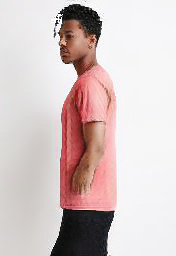}
\hspace{-4.5pt}
\includegraphics[width=0.108\textwidth]{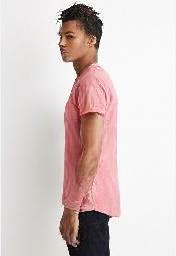}
\hspace{-4.5pt}
\includegraphics[width=0.108\textwidth]{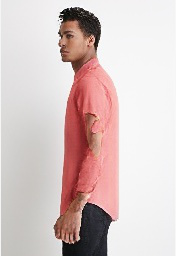}
\hspace{-4.5pt}
\includegraphics[width=0.108\textwidth]{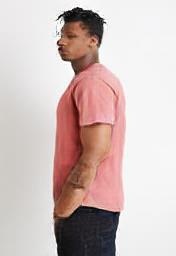}
\hspace{-4.5pt}
\includegraphics[width=0.108\textwidth]{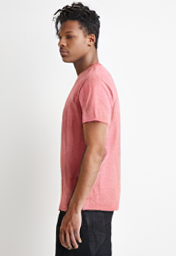}
\hspace{-4.5pt}
\includegraphics[width=0.108\textwidth]{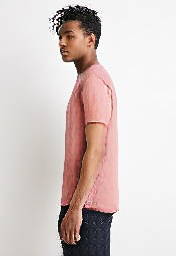}

\includegraphics[width=0.108\textwidth]{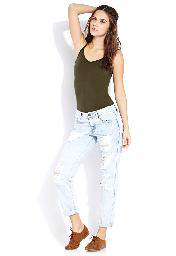}
\hspace{-4.5pt}
\includegraphics[width=0.108\textwidth]{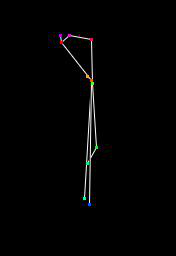}
\hspace{-4.5pt}
\includegraphics[width=0.108\textwidth]{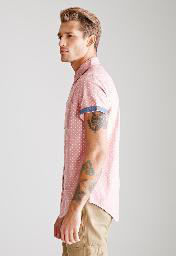}
\hspace{-4.5pt}
\includegraphics[width=0.108\textwidth]{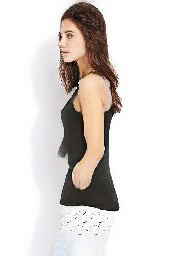}
\hspace{-4.5pt}
\includegraphics[width=0.108\textwidth]{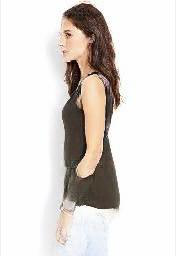}
\hspace{-4.5pt}
\includegraphics[width=0.108\textwidth]{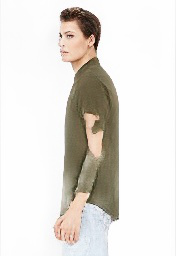}
\hspace{-4.5pt}
\includegraphics[width=0.108\textwidth]{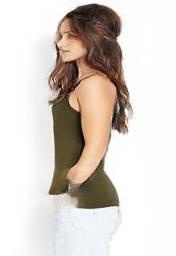}
\hspace{-4.5pt}
\includegraphics[width=0.108\textwidth]{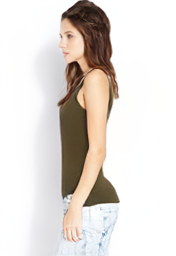}
\hspace{-4.5pt}
\includegraphics[width=0.108\textwidth]{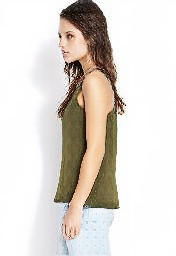}

\includegraphics[width=0.108\textwidth]{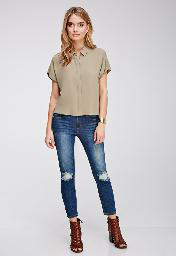}
\hspace{-4.5pt}
\includegraphics[width=0.108\textwidth]{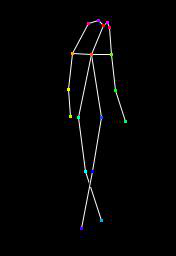}
\hspace{-4.5pt}
\includegraphics[width=0.108\textwidth]{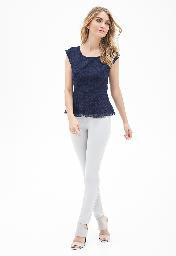}
\hspace{-4.5pt}
\includegraphics[width=0.108\textwidth]{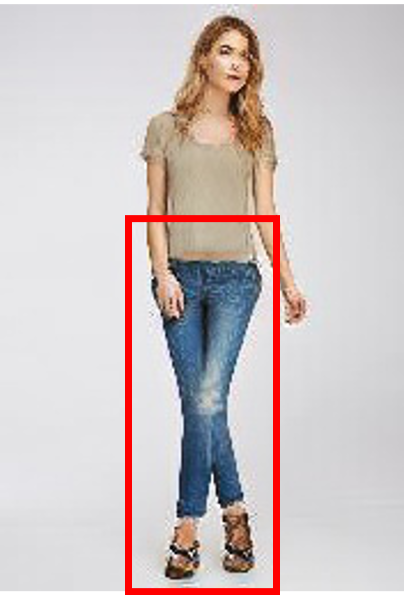}
\hspace{-4.5pt}
\includegraphics[width=0.108\textwidth]{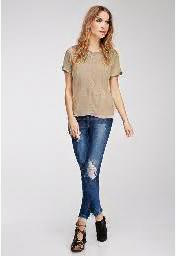}
\hspace{-4.5pt}
\includegraphics[width=0.108\textwidth]{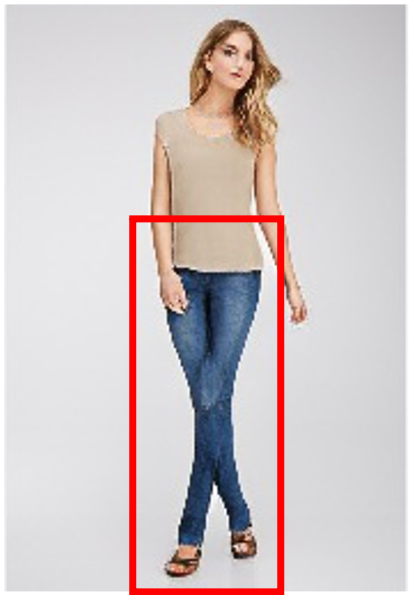}
\hspace{-4.5pt}
\includegraphics[width=0.108\textwidth]{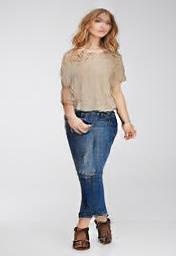}
\hspace{-4.5pt}
\includegraphics[width=0.108\textwidth]{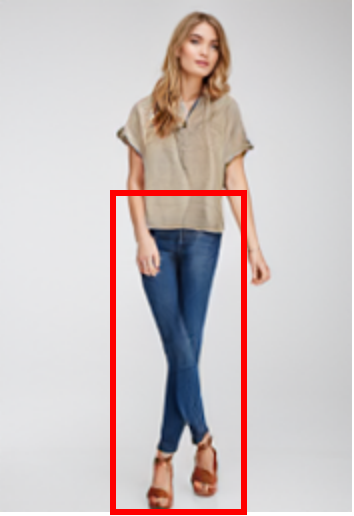}
\hspace{-4.5pt}
\includegraphics[width=0.108\textwidth]{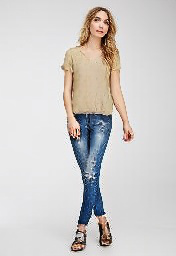}

\includegraphics[width=0.108\textwidth]{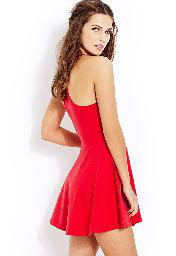}
\hspace{-4.5pt}
\includegraphics[width=0.108\textwidth]{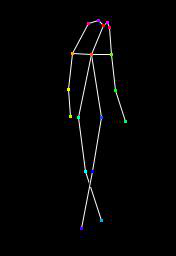}
\hspace{-4.5pt}
\includegraphics[width=0.108\textwidth]{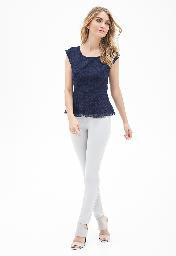}
\hspace{-4.5pt}
\includegraphics[width=0.108\textwidth]{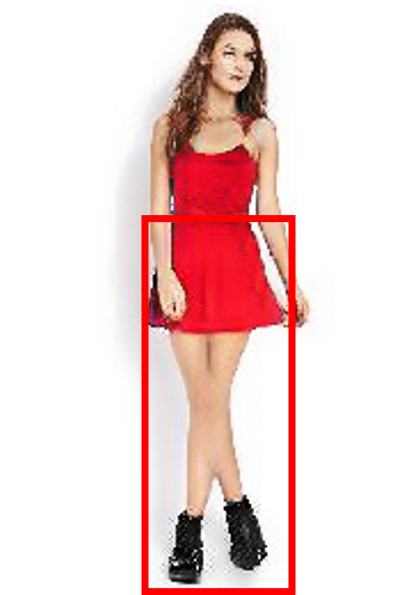}
\hspace{-4.5pt}
\includegraphics[width=0.108\textwidth]{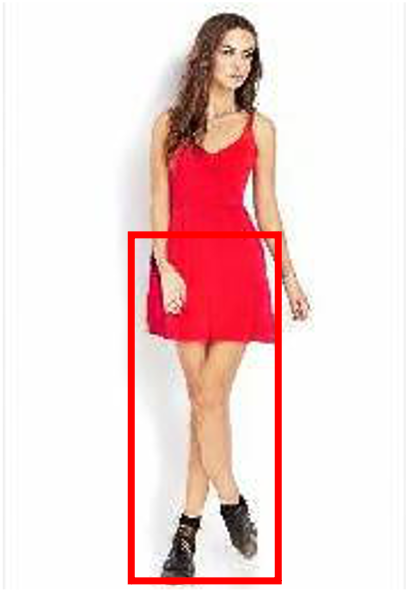}
\hspace{-4.5pt}
\includegraphics[width=0.108\textwidth]{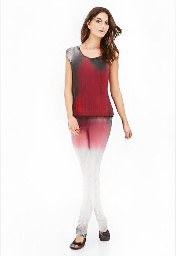}
\hspace{-4.5pt}
\includegraphics[width=0.108\textwidth]{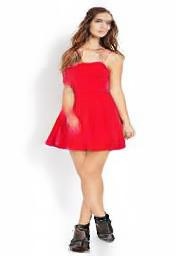}
\hspace{-4.5pt}
\includegraphics[width=0.108\textwidth]{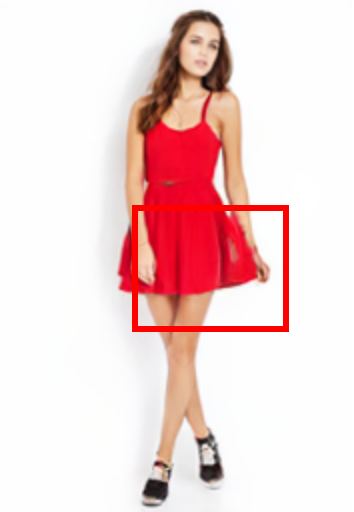}
\hspace{-4.5pt}
\includegraphics[width=0.108\textwidth]{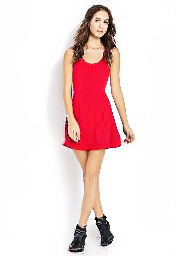}

\includegraphics[width=0.108\textwidth]{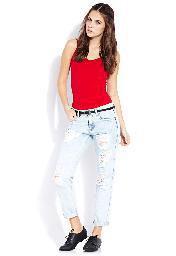}
\hspace{-4.5pt}
\includegraphics[width=0.108\textwidth]{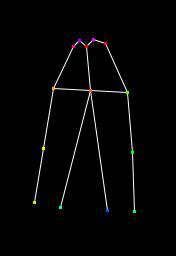}
\hspace{-4.5pt}
\includegraphics[width=0.108\textwidth]{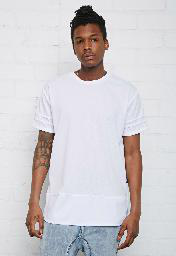}
\hspace{-4.5pt}
\includegraphics[width=0.108\textwidth]{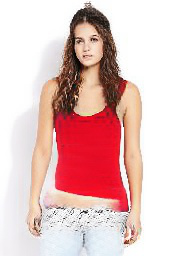}
\hspace{-4.5pt}
\includegraphics[width=0.108\textwidth]{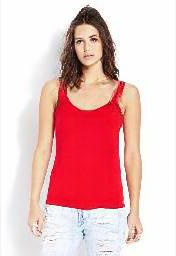}
\hspace{-4.5pt}
\includegraphics[width=0.108\textwidth]{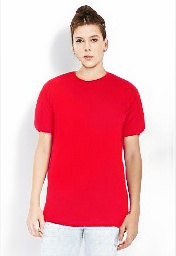}
\hspace{-4.5pt}
\includegraphics[width=0.108\textwidth]{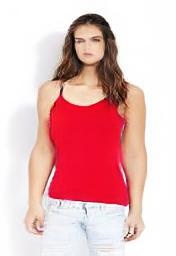}
\hspace{-4.5pt}
\includegraphics[width=0.108\textwidth]{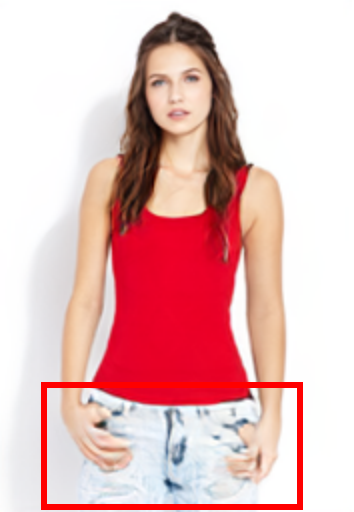}
\hspace{-4.5pt}
\includegraphics[width=0.108\textwidth]{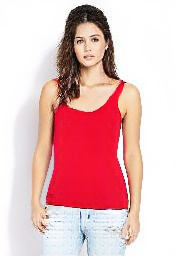}

\includegraphics[width=0.108\textwidth]{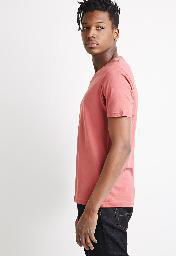}
\hspace{-4.5pt}
\includegraphics[width=0.108\textwidth]{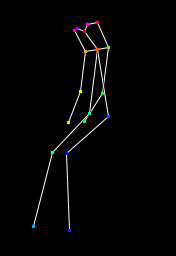}
\hspace{-4.5pt}
\includegraphics[width=0.108\textwidth]{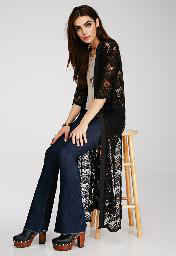}
\hspace{-4.5pt}
\includegraphics[width=0.108\textwidth]{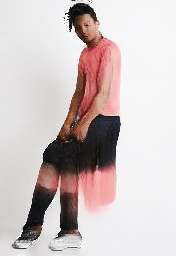}
\hspace{-4.5pt}
\includegraphics[width=0.108\textwidth]{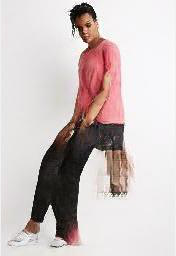}
\hspace{-4.5pt}
\includegraphics[width=0.108\textwidth]{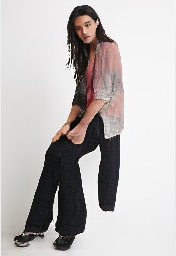}
\hspace{-4.5pt}
\includegraphics[width=0.108\textwidth]{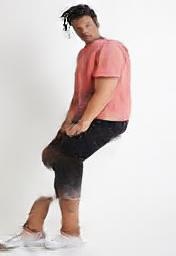}
\hspace{-4.5pt}
\includegraphics[width=0.108\textwidth]{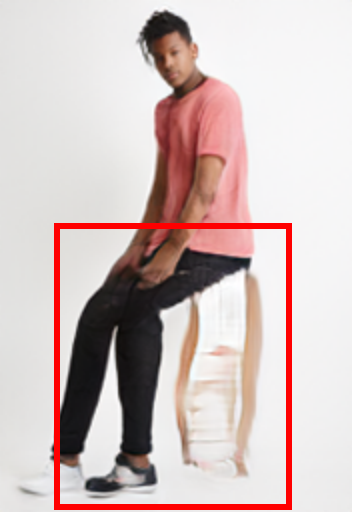}
\hspace{-4.5pt}
\includegraphics[width=0.108\textwidth]{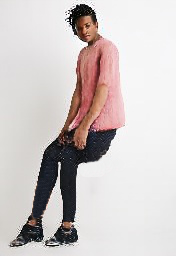}

\begin{subfigure}{\textwidth}
  \caption[l]{ \hspace{ 16pt}  Person \hspace{ 26pt} Pose  \hspace{ 28pt} Source\hspace{ 28pt} ADGAN \hspace{ 26pt} PISE\hspace{ 30pt} SPGNet \hspace{ 26pt}  CASD \hspace{ 28pt}  NTED \hspace{ 26pt} Ours}
\end{subfigure}

\caption{Performance comparison on swap pose transfer task.}
\label{fig:end_comp_swap_pose}
\vspace{-0.3cm}
\end{figure*}

We shall emphasize that although the quantitative improvement in pose transfer performance of the state-of-the-art methods SPG~\cite{lv2021learning} and PISE~\cite{zhang2021pise} is marginal, the improvements in swap pose transfer (Figure \ref{fig:end_comp_swap_pose}) and attribute transfer tasks are obvious. Our method produces person images maintaining the attributes consistent to the source/target person, while PISE and SPG struggle to generate a person of the same gender as the source person (Figure~\ref{fig:swap_pose}) or maintain the style consistency (Figure~\ref{fig:attribute_transfer}). Although NTED~\cite{ren2022neural} produces the best FID and LIPIPS scores with the fact that it takes a coarse-to-fine generation strategy to deploy the NTED operations at different scales, in the second row of Figure \ref{fig:pose} a bag that does not exist in the input image was created by NTED along with the female model. The similar phenomenon can also be observed in the $8_{th}$ column of Figure~\ref{fig:swap_pose}, where NTED additionally generated a chair. These reflect that NTED fails to completely disentangle the representation as it opts to introduce nonnegligible residues of the target image into the generated output. By contrast, due to the well-learned latent representation our method performs pose transfer with consistent semantic regions without being affected by artifacts.

\begin{table}[t!]
\footnotesize
\centering
\caption{ User study of different methods on person image generation tasks. N/ indicates that the method is not able to perform on this task.}
\begin{tabular}{l|c|c|c}
\hlineB{2} 
\multirow{2}{*}{Setting}&\multicolumn{3}{c}{User study $\uparrow$}\\
\cline{2-4}
 &Component attribute transfer &Pose transfer &Swap pose transfer\\
\hline 
 ADGAN  &32.63\%& 20.75\%&21.2\%\\  
 PISE  &29.01\%& 24.34\% &27.6\% \\
 SPG    & N/   & 28.28\%   &13.4\% \\
 Ours  &38.36\%& 26.63\% &37.8\% \\ 
\hlineB{2} 
\end{tabular}
\label{tab:user}
\end{table}

\subsubsection{Market1501 Dataset} 
We also evaluate our model on the Market1501 dataset. Note that SPG~\cite{lv2021learning} is the state-of-the-art method designed for pose transfer only without learning representations for attribute transfer, while ADGAN~\cite{men2020controllable}, PISE~\cite{zhang2021pise} and our method work for both pose transfer and attribute transfer. In spite of this, our method achieves comparable performance to SPG. See Table~\ref{tab:exp_Market1501} and Figure~\ref{fig:exp_Market1501}. These strongly demonstrate the superiority of our method.


\subsection{Component Attribute Transfer}
\begin{figure}[t!]
\captionsetup[subfigure]{justification=raggedright, singlelinecheck=false, labelformat=empty}    
 \centering
 \raisebox{0.03\textwidth }{\makebox[0.11\textwidth]{\parbox{0.01\linewidth}{}}}
\hspace{-4pt}
\includegraphics[width=0.11\textwidth]{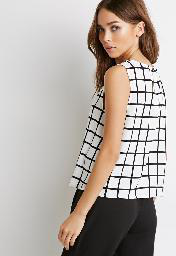}
\hspace{-4pt}
\includegraphics[width=0.11\textwidth]{fig_sup/cross1/0}
\hspace{-4pt}
\includegraphics[width=0.11\textwidth]{fig_sup/cross1/0}
\raisebox{0.03\textwidth }{\makebox[0.11\textwidth]{\parbox{0.01\linewidth}{}}}
\hspace{-4pt}
\includegraphics[width=0.11\textwidth]{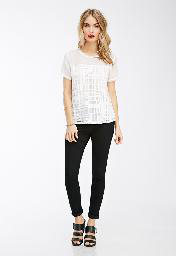}
\hspace{-4pt}
\includegraphics[width=0.11\textwidth]{fig_sup/cross2/0}
\hspace{-4pt}
\includegraphics[width=0.11\textwidth]{fig_sup/cross2/0}\\
 \includegraphics[width=0.11\textwidth]{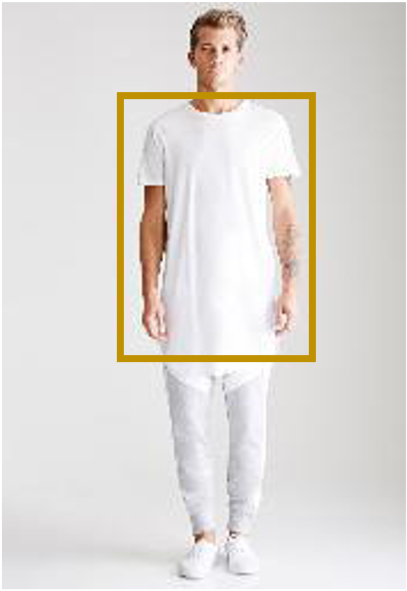}
\hspace{-4pt}  
\includegraphics[width=0.11\textwidth]{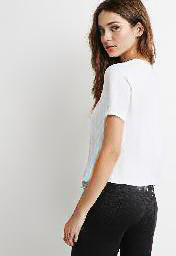}
\hspace{-4pt}
\includegraphics[width=0.11\textwidth]{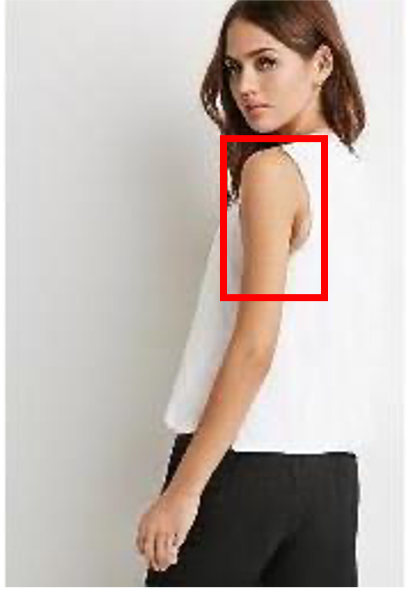}
\hspace{-4pt}
\includegraphics[width=0.11\textwidth]{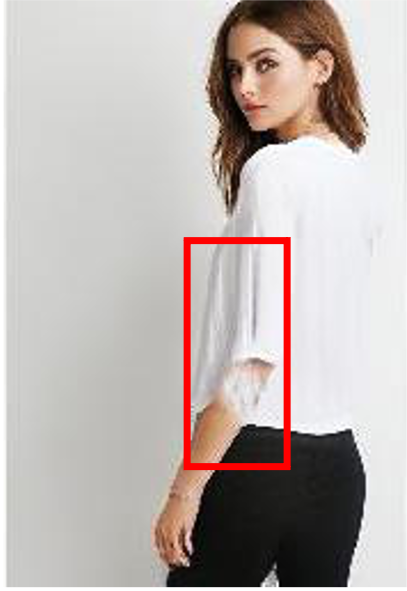}
 \includegraphics[width=0.11\textwidth]{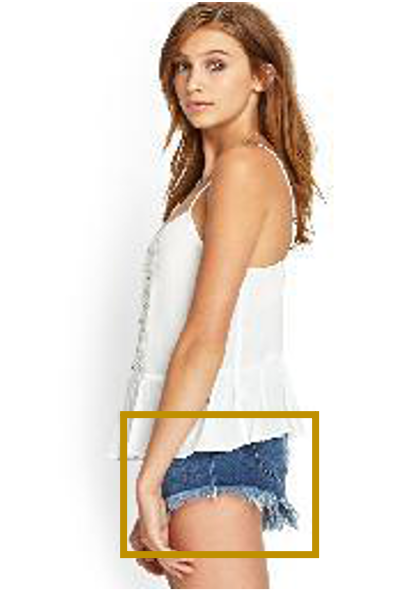}
\hspace{-4pt}
\includegraphics[width=0.11\textwidth]{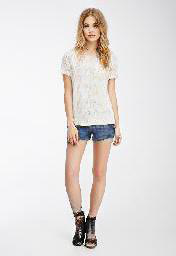}
\hspace{-4pt} 
\includegraphics[width=0.11\textwidth]{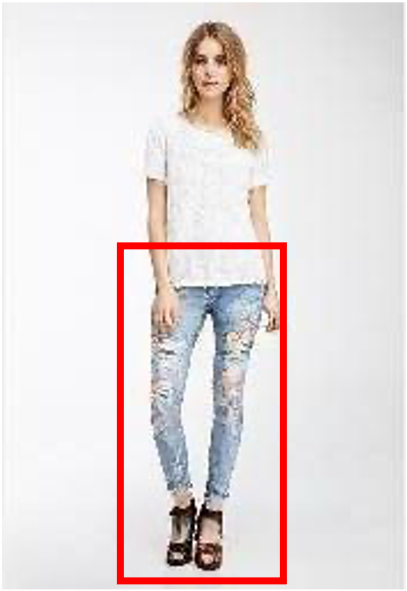}
\hspace{-4pt}
\includegraphics[width=0.11\textwidth]{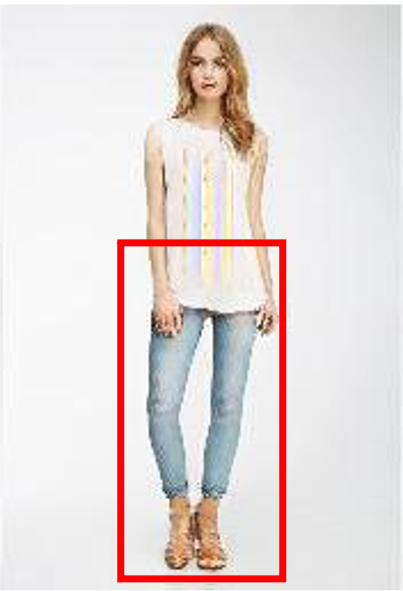}\\
 \includegraphics[width=0.11\textwidth]{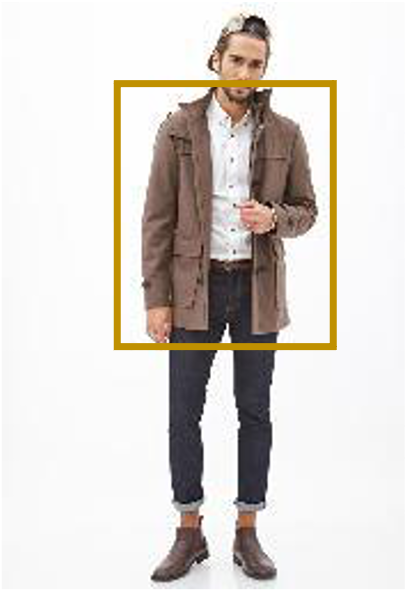}
\hspace{-4pt}  
\includegraphics[width=0.11\textwidth]{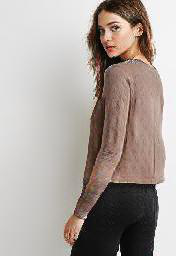}
\hspace{-4pt}
\includegraphics[width=0.11\textwidth]{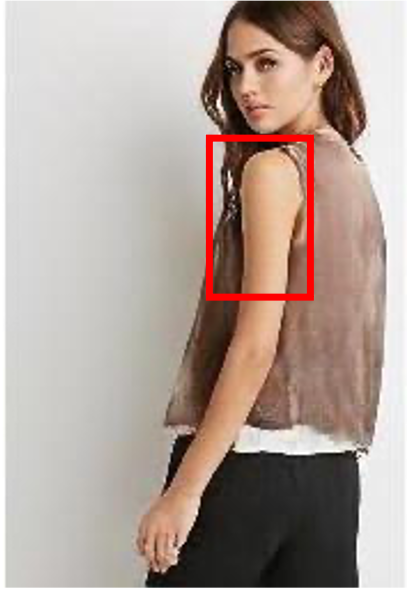}
\hspace{-4pt}
\includegraphics[width=0.11\textwidth]{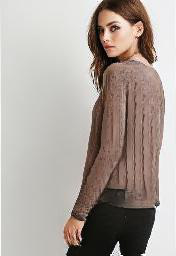}
 \includegraphics[width=0.11\textwidth]{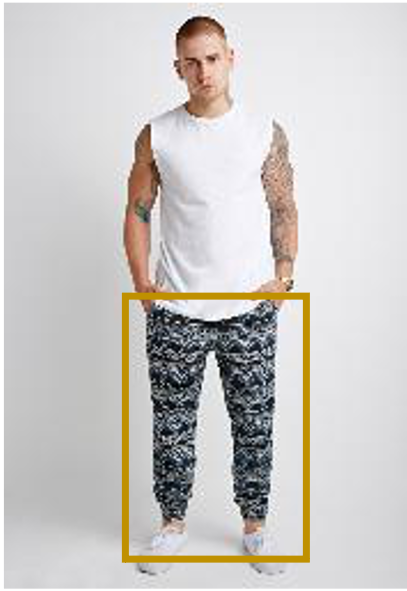}
\hspace{-4pt} 
\includegraphics[width=0.11\textwidth]{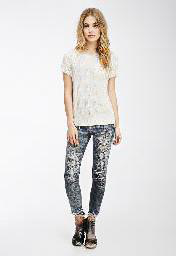}
\hspace{-4pt}
\includegraphics[width=0.11\textwidth]{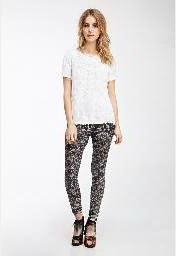}
\hspace{-4pt}
\includegraphics[width=0.11\textwidth]{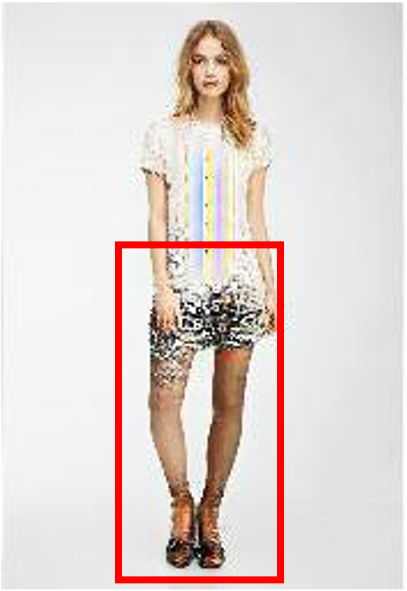}\\
 \begin{subfigure}{\textwidth}
  \caption[l]{ \hspace{ 8pt}  Cloth   \hspace{ 6pt} Ours\hspace{ 8pt} PISE  \hspace{ 0pt} ADGAN \hspace{ 2pt} Pant  \hspace{ 8pt}Ours \hspace{ 4pt} PISE \hspace{ 2pt}ADGAN}
    \end{subfigure}
\caption{Component attribute transfer results. }\label{fig:attribute_transfer}
\end{figure}

Component attribute transfer is the replacement of the attribute of a person in the source image with that of another person in the target image while preserving other attributes of the source person. 
We compare our proposed DRL-CPG with ADGAN and PISE which are able to edit component-level human attributes based on the corresponding semantic mask that provides guidance to separate each component at the image level, and summarize the results in Figure \ref{fig:attribute_transfer}.  As we can observe, our DRL-CPG generates natural images with new attributes introduced harmoniously while preserving the textures of the remaining components. 
In contrast, ADGAN fails to create a consistent sleeve style. This can be explained by the fact that the representations learned by ADGAN are not well disentangled; PISE generates person images with new attributes but fails to change the shape of the clothes accordingly.
 
\subsection{User Study}
We report the results of a user study comparing
our model to ADGAN, PISE and SPG on pose transfer,
component attribute transfer, and swap pose transfer tasks. For the tasks, we select 200 pairs from the
test subset to synthesize person images. Eight participants are required to make a choice about which output they prefer in terms of the visual quality, and more importantly, the pose and component attribute consistency with ground-truth. Note that SPG does not learn disentangled latent representation for different attribute components. This is partly the reason that it works well on pose transfer tasks but fails to work on tasks requiring disentanglement like swap pose transfer. Inherently, it does not hold the capability of conducting component attribute transfer tasks.

\begin{figure}[t!]
\captionsetup[subfigure]{justification=raggedright, singlelinecheck=false, labelformat=empty}    
 \centering
    \begin{subfigure}{\linewidth}
     \centering
\includegraphics[width=0.16\textwidth]{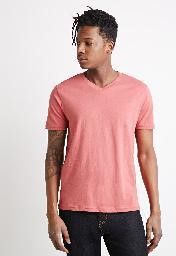}
\hspace{-4.5pt}
\includegraphics[width=0.16\textwidth]{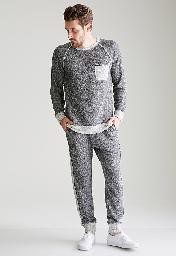}
\hspace{-4.5pt}
\includegraphics[width=0.16\textwidth]{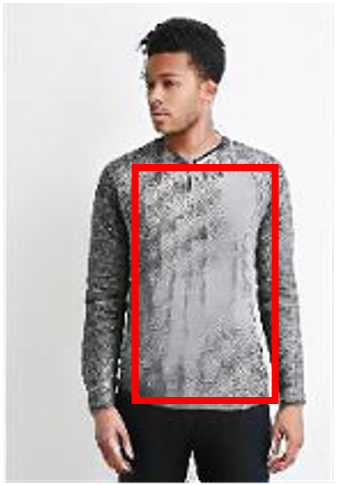}
\hspace{-4.5pt}
\includegraphics[width=0.16\textwidth]{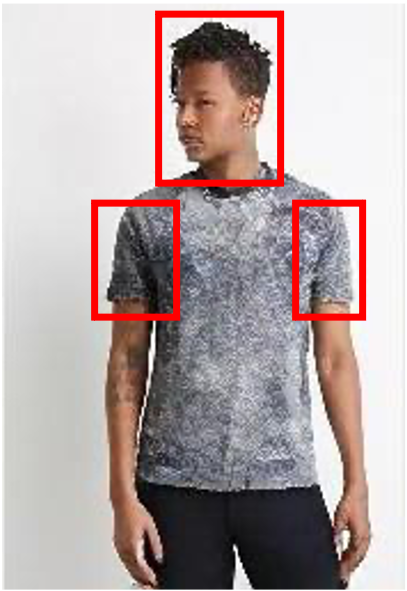}
\hspace{-4.5pt}
\includegraphics[width=0.16\textwidth]{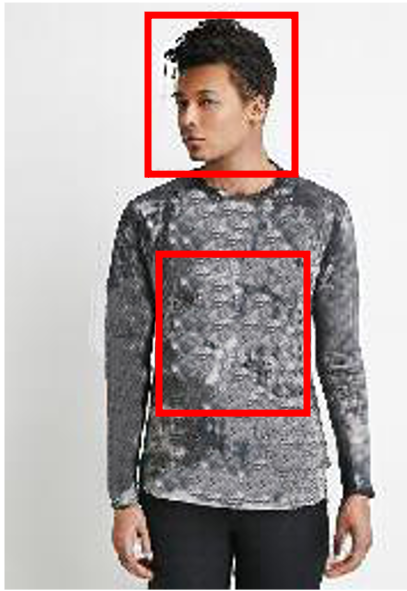}
\hspace{-4.5pt}
\includegraphics[width=0.16\textwidth]{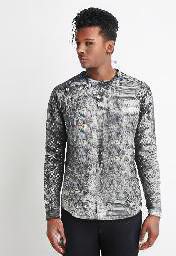}
\hspace{-4.5pt}

\includegraphics[width=0.16\textwidth]{posecomp/cross}
\hspace{-4.5pt}
\includegraphics[width=0.16\textwidth]{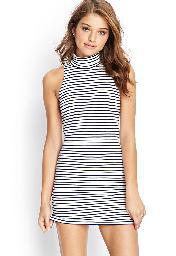}
\hspace{-4.5pt}
\includegraphics[width=0.16\textwidth]{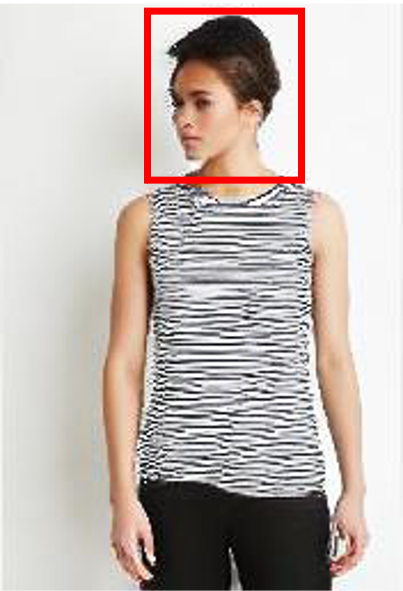}
\hspace{-4.5pt}
\includegraphics[width=0.16\textwidth]{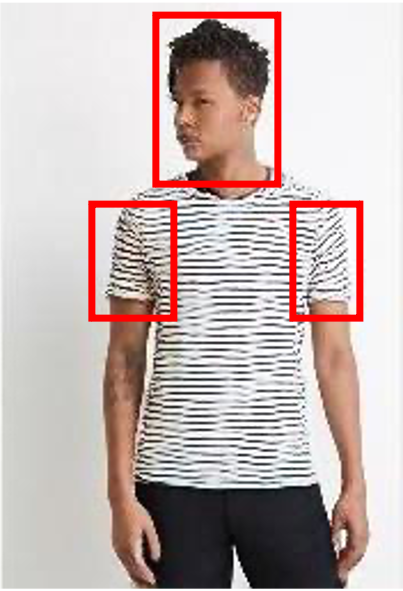}
\hspace{-4.5pt}
\includegraphics[width=0.16\textwidth]{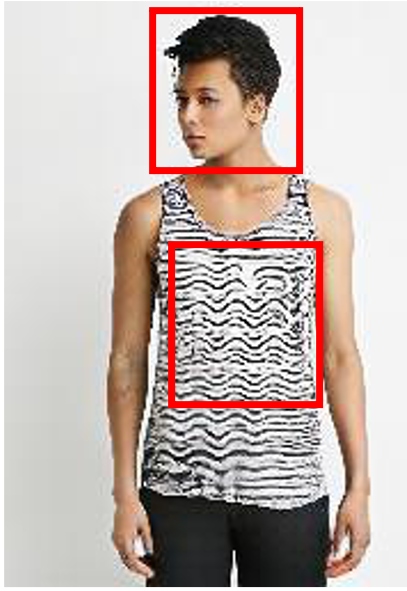}
\hspace{-4.5pt}
\includegraphics[width=0.16\textwidth]{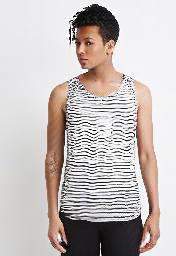}
\hspace{-4.5pt}

  \caption[l]{  \hspace{ 8pt} Source \hspace{ 8pt} Target\hspace{ 10pt} ADGAN\hspace{ 16pt} PISE\hspace{ 8pt} Base+MA \hspace{ 10pt} Ours }
     \vspace{-0.07cm}
     
\end{subfigure}
\caption{Performance comparison in terms of fidelity.}
\label{fig:end_comp}
\end{figure}

Although the quantitative improvement in pose transfer is marginal, the improvements in swap pose transfer and attribute transfer tasks are obvious. Our method produces person images maintaining the attributes consistent to the source/target person, while PISE and SPG struggle to generate person of the same gender as the source person and maintain the style consistency (Figure \ref{fig:end_comp}). Results are listed in Table \ref{tab:user}. We can see the results of ours outperform the state of the arts, which further indicates the superiority of our method.

\setlength{\tabcolsep}{6pt}
\begin{table}[t!]
\caption{Ablation study on pose transfer task.}
\centering
\footnotesize
\begin{tabular}{l|c|c|c}
\hlineB{2} 
 Settings& FID$\downarrow$  & SSIM$\uparrow$ &LPIPS $\downarrow$ \\
\hline
Base  &38.238&0.362&0.3537 \\ 
Base + MA  & 15.521 &0.772&0.229 \\ 
Base + MA + CL & 14.845&0.781&0.217 \\
Base + MA + CL + Mask (DRL-CPG)&{\color{red} 13.514}&{\color{red} 0.792}&{\color{red} 0.2027}\\
\hlineB{2} 
\end{tabular}
\label{tab:Ablation}
\end{table}

\subsection{Ablation Study}
We ablate our training mechanism by training our model with different combinations of strategies. To evaluate the effectiveness of the training strategies assisting in learning disentangled representations, we test the trained model variants without masks. In this way, we feed the complete person image into the trained model for person generation tasks. Note our full model is tested with masks to extract the components.

\bfsection{Base model} We train a base model without any strategy, and test the trained model without masks. 

\bfsection{Base + MA model} The random component mask-agnostic (MA) strategy is included to train the base model. 

\bfsection{Base + MA + CL model} The curriculum learning (CL) strategy is further added to improve performance.

\bfsection{Base + MA + CL + Mask (DRL-CPG) model} Given the previous model variants are tested without masks, we take all the masks to extract components at the testing stage. This is the proposed full model.

\begin{figure}[t!]
\captionsetup[subfigure]{justification=raggedright, singlelinecheck=false, labelformat=empty}    
 \centering
\includegraphics[width=0.12\textwidth]{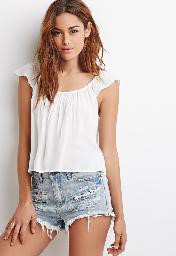}
\hspace{-5pt}
\includegraphics[width=0.12\textwidth]{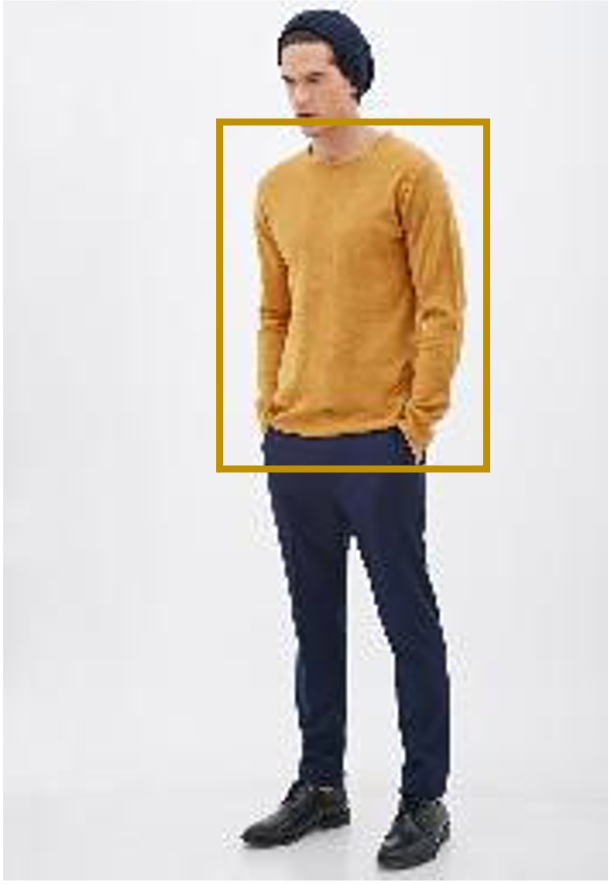}
\hspace{-5pt}
\includegraphics[width=0.12\textwidth]{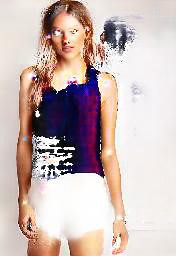}
\hspace{-5pt}
\includegraphics[width=0.12\textwidth]{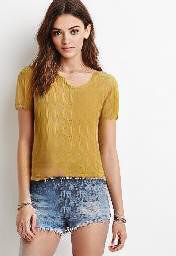}
\hspace{-5pt}
\includegraphics[width=0.12\textwidth]{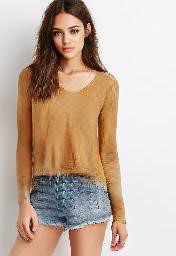}
\hspace{-5pt}
\includegraphics[width=0.12\textwidth]{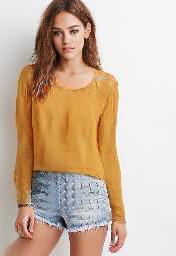}
\hspace{-5pt}
\includegraphics[width=0.12\textwidth]{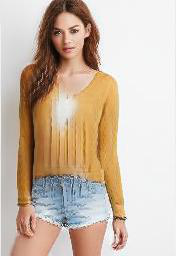}
\hspace{-5pt}
\includegraphics[width=0.12\textwidth]{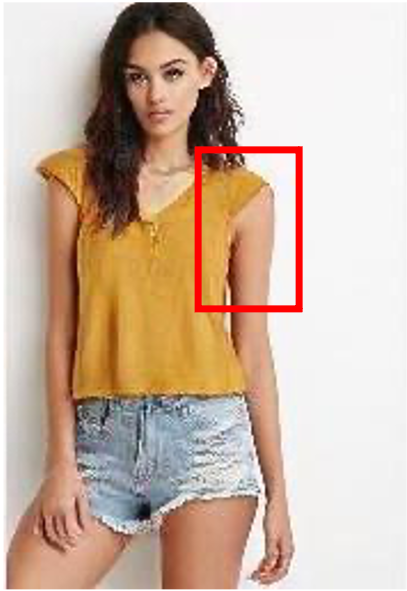}\\
\includegraphics[width=0.12\textwidth]{crosscomp/cross}
\hspace{-5pt}
\includegraphics[width=0.12\textwidth]{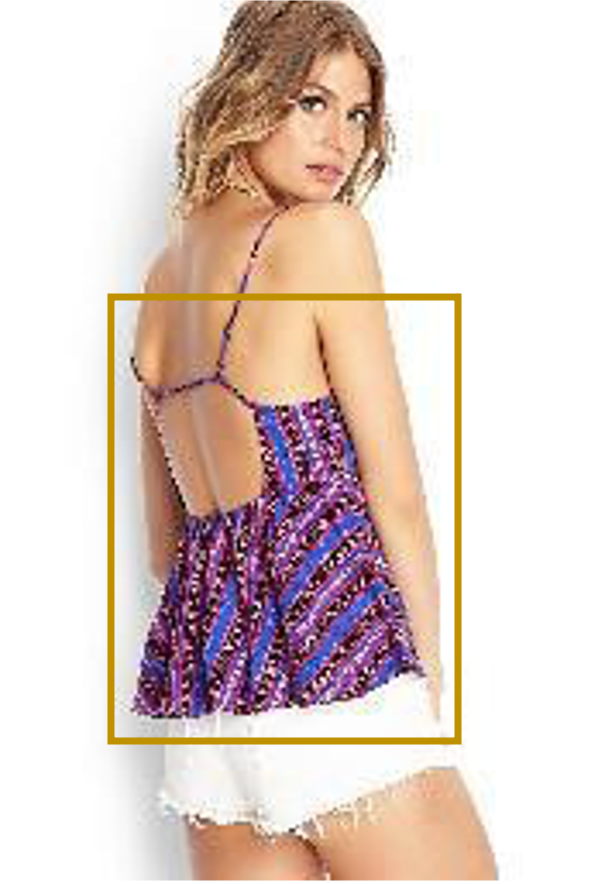}
\hspace{-5pt}
\includegraphics[width=0.12\textwidth]{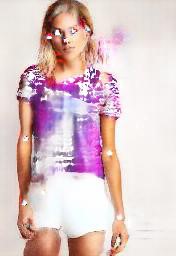}
\hspace{-5pt}
\includegraphics[width=0.12\textwidth]{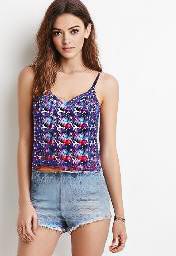}
\hspace{-5pt}
\includegraphics[width=0.12\textwidth]{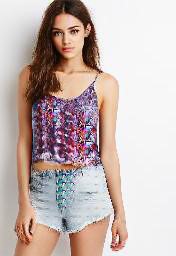}
\hspace{-5pt}
\includegraphics[width=0.12\textwidth]{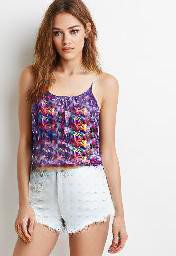}
\hspace{-5pt}
\includegraphics[width=0.12\textwidth]{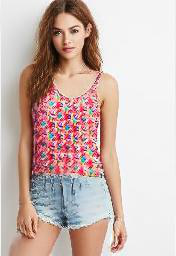}
\hspace{-5pt}
\includegraphics[width=0.12\textwidth]{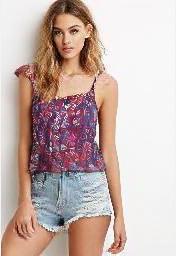}\\

  \begin{subfigure}{\textwidth}
  \caption[l]{ \hspace{ 1pt}  Person   \hspace{ 4pt} Cloth\hspace{ 4pt} Base  \hspace{ 4pt} +MA \hspace{ 2pt} +MA+CL  \hspace{ 4pt}Ours \hspace{ 0pt} ADGAN \hspace{ 2pt}PISE}
    \end{subfigure}

\caption{Ablation study on component attribute transfer task. The results of ADGAN and PISE are listed as references.}
\label{fig:Ablation}
\end{figure}


\begin{figure}[t!]
\captionsetup[subfigure]{justification=raggedright, singlelinecheck=false, labelformat=empty}    
 \centering
    \begin{subfigure}{\linewidth}
     \centering
\includegraphics[width=0.16\textwidth]{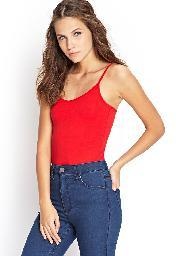}
\hspace{-4.5pt}
\includegraphics[width=0.16\textwidth]{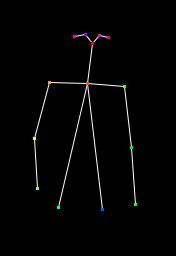}
\hspace{-4.5pt}
\includegraphics[width=0.16\textwidth]{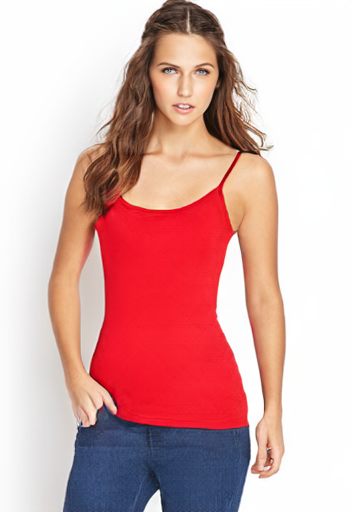}
\hspace{-4.5pt}
\includegraphics[width=0.16\textwidth]{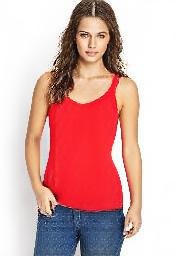}
\hspace{-4.5pt}
\includegraphics[width=0.16\textwidth]{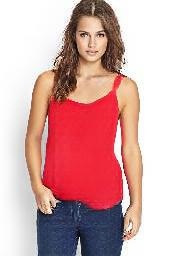}
\hspace{-4.5pt}
\includegraphics[width=0.16\textwidth]{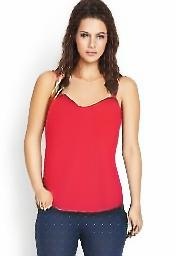}
\hspace{-4.5pt}

\includegraphics[width=0.16\textwidth]{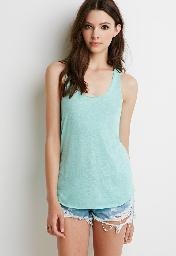}
\hspace{-4.5pt}
\includegraphics[width=0.16\textwidth]{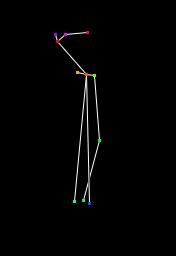}
\hspace{-4.5pt}
\includegraphics[width=0.16\textwidth]{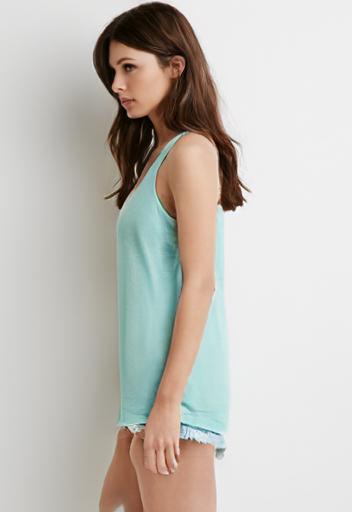}
\hspace{-4.5pt}
\includegraphics[width=0.16\textwidth]{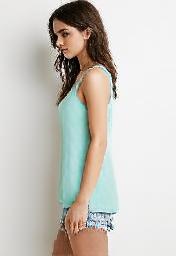}
\hspace{-4.5pt}
\includegraphics[width=0.16\textwidth]{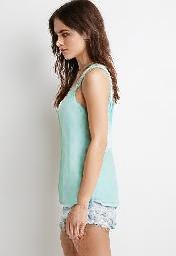}
\hspace{-4.5pt}
\includegraphics[width=0.16\textwidth]{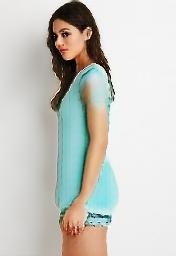}
\hspace{-4.5pt}

  \caption[l]{  \hspace{ 8pt} Source \hspace{ 8pt} Target\hspace{ 8pt} Ours@{\color{blue}512}\hspace{ 6pt} Ours@{\color{blue}256}\hspace{ 4pt} w/o $\lambda_{ctx}$\hspace{ 4pt} w/o $\mathcal{L}_{per}$}
     \vspace{-0.07cm}

\end{subfigure}
\caption{Ablation study on loss function and resolution.}
\label{fig:loss_function}
\end{figure}

\setlength{\tabcolsep}{10pt}
 \begin{table}[t!]
\centering
\vspace{-0.2cm}
\caption{Performance of Transformer (TX) at different sizes.}\label{tab:transformer_different_size}
\begin{tabular}{l|c|c|c}
\hlineB{2} 
 Settings& FID$\downarrow$  & SSIM$\uparrow$ &LPIPS $\downarrow$ \\
\hline  
No-TX &18.374&0.762&0.2403  \\ 
\hline
Small-TX &16.386  &0.775&0.2297 \\ 
Medium-TX &{\color{red} 13.514}&{\color{red} 0.792}&{\color{red} 0.2027}\\
Large-TX &14.403  &0.777&0.2170 \\
\hlineB{2} 
\end{tabular}
\vspace{-0.3cm}
\end{table}
\setlength{\tabcolsep}{10pt}
 \begin{table}[t!]
\centering
\vspace{-0.2cm}
\caption{Ablation study on different loss functions.}\label{tab:ablation_loss}
\begin{tabular}{l|c|c|c}
\hlineB{2} 
 Settings& FID$\downarrow$  & SSIM$\uparrow$ &LPIPS $\downarrow$ \\
\hline  
Full model &{\color{red} 13.514}&{\color{red} 0.792}&{\color{red} 0.2027} \\ 
w/o $\lambda_{ctx}$ &13.724  &0.787&0.2070 \\
w/o $\mathcal{L}_{per}$&13.962  &0.772&0.2121 \\
\hlineB{2} 
\end{tabular}
\vspace{-0.3cm}
\end{table}

We summarize the quantitative results in Table \ref{tab:Ablation} and demonstrate the qualitative comparison in Figure \ref{fig:Ablation}. 
At the testing stage, the base model fails to generate realistic images without masks. After taking the random component mask-agnostic strategy to train the model, the trained model is able to generate images without masks at the testing stage. This shows the capability of the MA strategy in assisting the transformer encoder to recognize each component and extract meaningful representation from the complete person images. The curriculum learning strategy further improves the performance. Our full model generates the best person images by taking full use of segmentation masks at the testing stage.

\begin{figure}[t!]
\begin{subfigure}{0.12\linewidth}
        \centering
        \includegraphics[width=\textwidth]{fig_sup/pose3/0}
    \end{subfigure} 
\rule[-28px]{0.8px}{60px}
\begin{minipage}{0.8\linewidth}
    \begin{subfigure}{\linewidth}
\includegraphics[width=0.17\textwidth]{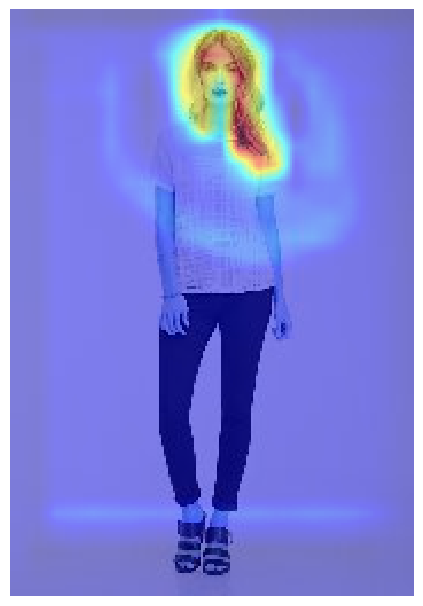}
\hspace{-6pt}
\includegraphics[width=0.17\textwidth]{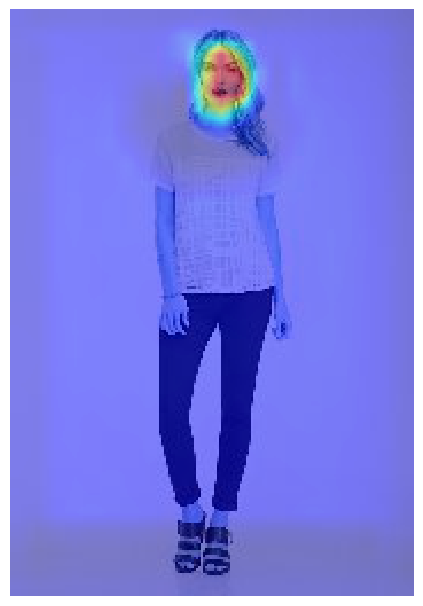}
\hspace{-6pt}
\includegraphics[width=0.17\textwidth]{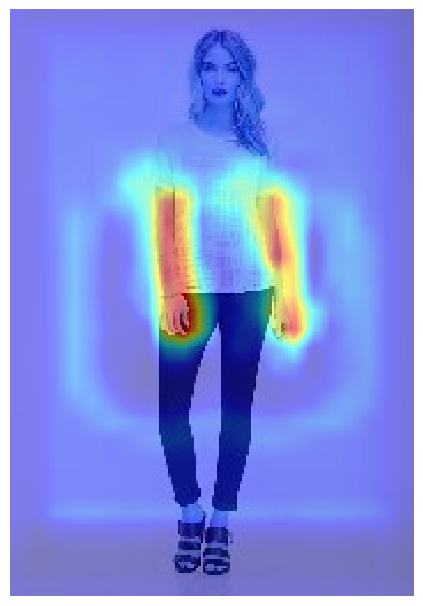}
\hspace{-6pt}
\includegraphics[width=0.17\textwidth]{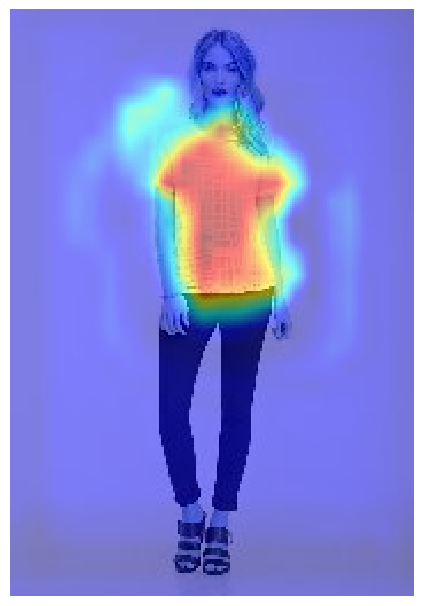}
\hspace{-6pt}
\includegraphics[width=0.17\textwidth]{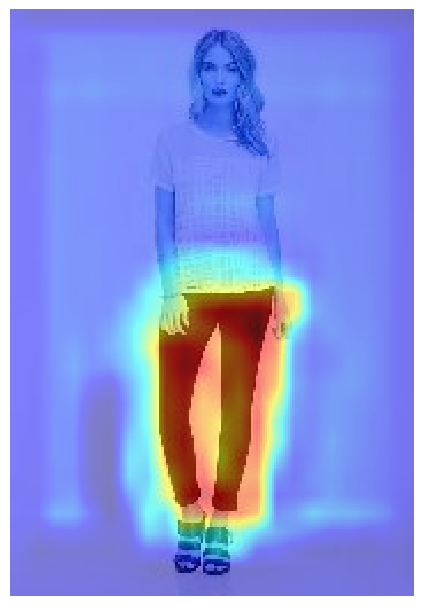}
\hspace{-6pt}
\includegraphics[width=0.17\textwidth]{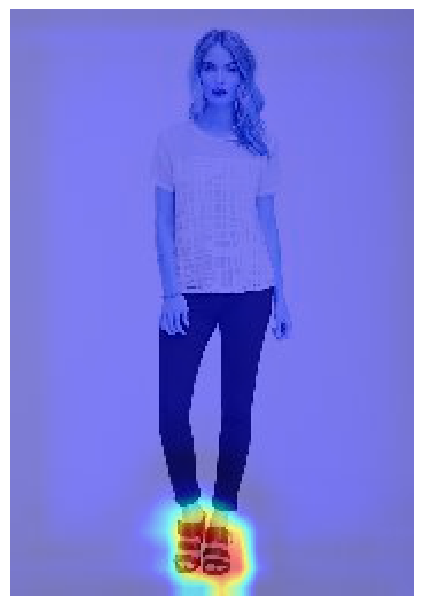}
    \end{subfigure}

    \vspace{-2pt}

    \begin{subfigure}{\linewidth}
\includegraphics[width=0.17\textwidth]{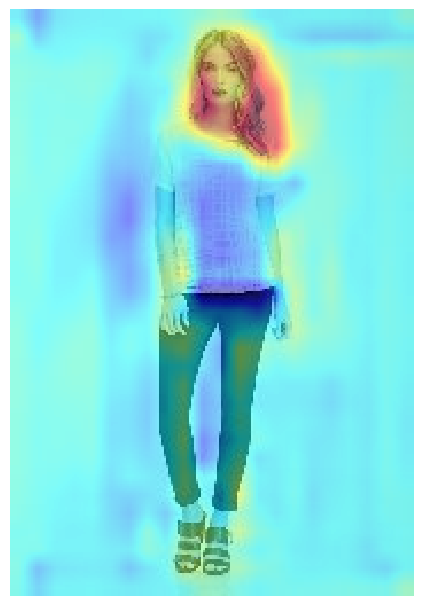}
\hspace{-6pt}
\includegraphics[width=0.17\textwidth]{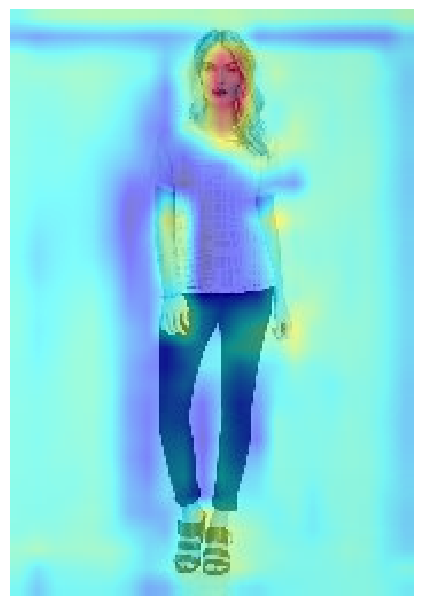}
\hspace{-6pt}
\includegraphics[width=0.17\textwidth]{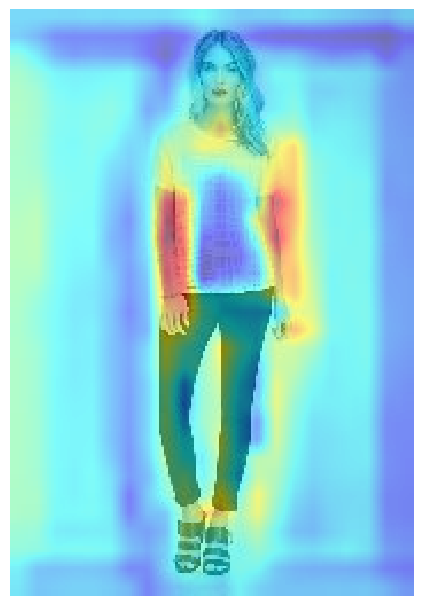}
\hspace{-6pt}
\includegraphics[width=0.17\textwidth]{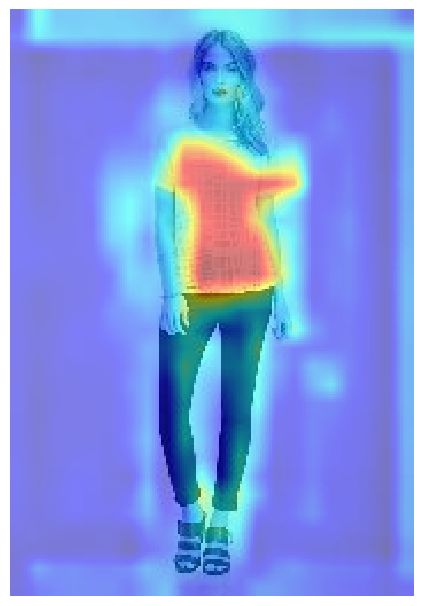}
\hspace{-6pt}
\includegraphics[width=0.17\textwidth]{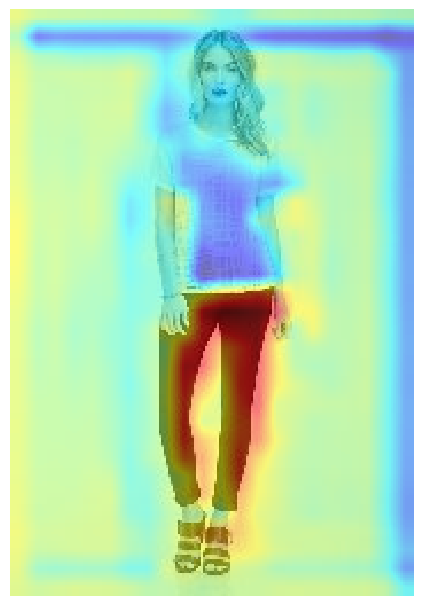}
\hspace{-6pt}
\includegraphics[width=0.17\textwidth]{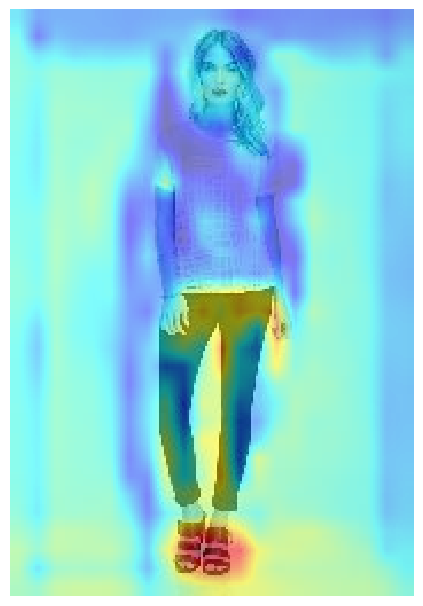}
    \end{subfigure}
    
    \vspace{-2pt}

    \begin{subfigure}{\linewidth}
\includegraphics[width=0.17\textwidth]{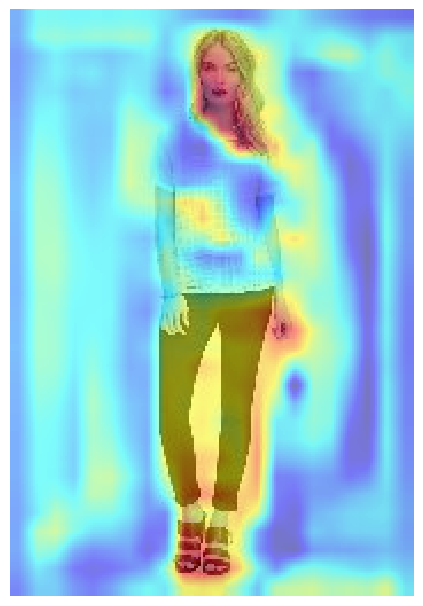}
\hspace{-6pt}
\includegraphics[width=0.17\textwidth]{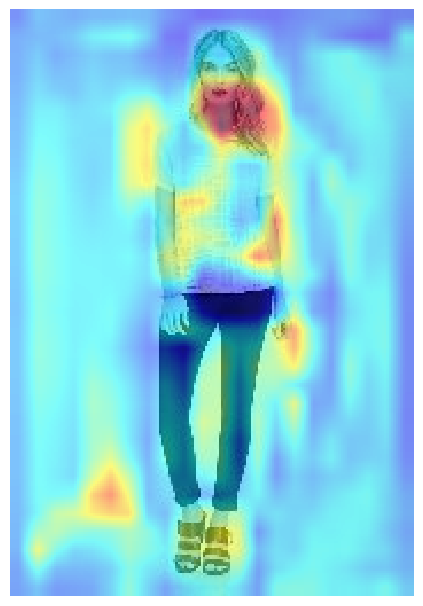}
\hspace{-6pt}
\includegraphics[width=0.17\textwidth]{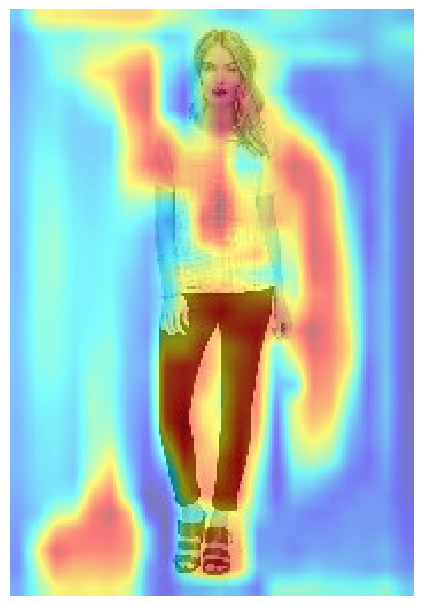}
\hspace{-6pt}
\includegraphics[width=0.17\textwidth]{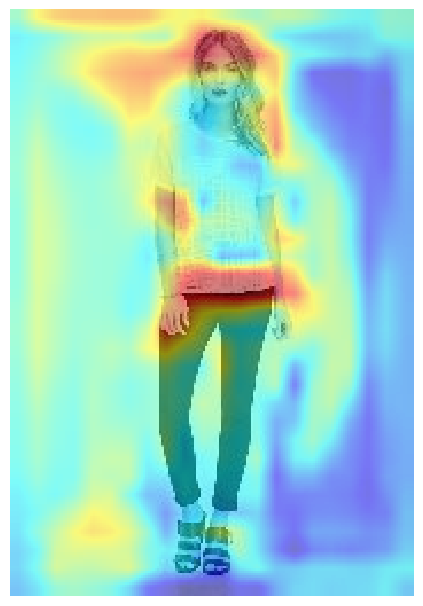}
\hspace{-6pt}
\includegraphics[width=0.17\textwidth]{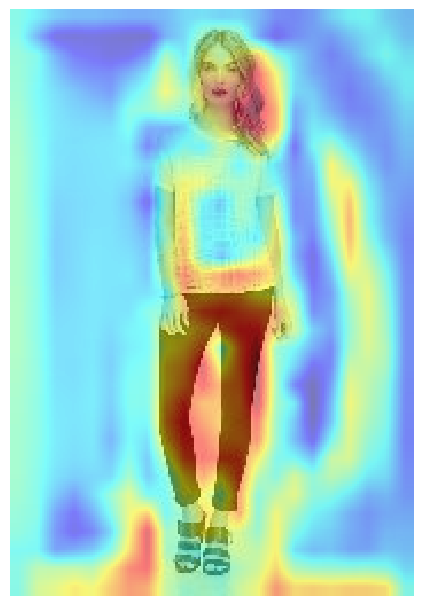}
\hspace{-6pt}
\includegraphics[width=0.17\textwidth]{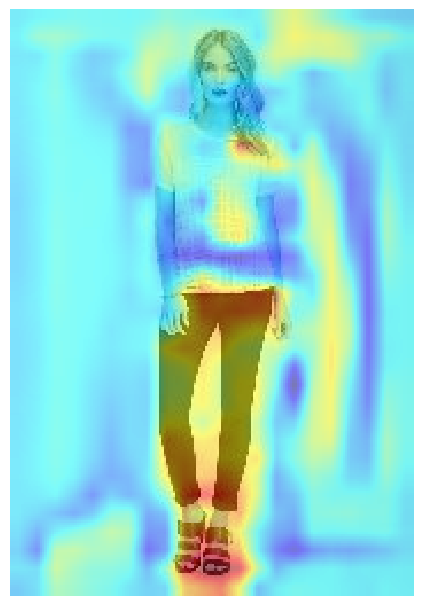}
    \end{subfigure}    
\end{minipage}%
\caption{Visualize of attribute distribution map of $\text{Base + MA + CL + Mask (DRL-CPG) model}$ (top), $\text{Base + MA + CL model}$ (middle) and $\text{Base model}$  (bottom).}\label{fig:dec_att}
\vspace{-0.3cm}
    \end{figure}

\bfsection{Resolution} We train another model with minor modifications on images of $512 \times 352$ resolution. As shown in Figure \ref{fig:loss_function}, we observe that our model@512 produces faithful synthetic images with well-preserved texture details. It demonstrates that training on high-resolution images consistently learns the model to better distinguish boundaries between different semantic regions and significantly improves the generation quality.

\bfsection{Loss function} We further compare the effect of different loss in Figure \ref{fig:loss_function} and list the quantitative scores in Table \ref{tab:ablation_loss}. As can be seen, w/o contextual loss $\lambda_{ctx}$ the model fails to achieve the completeness of semantic regions, such as the cloth in the first row of Figure \ref{fig:loss_function}. Without making use of the perceptual loss $\mathcal{L}_{per}$, the model tends to create artifacts in the synthesized image, making the generation quality degraded to some degree.

\subsection{Analysis and Discussion}
We conduct further experiments for analysis and discussion of the effectiveness of transformers and disentangling latent representation in the attribute encoder.

\bfsection{Effectiveness of preserving texture details}
We show the comparison on the fidelity of cloth texture in Figure \ref{fig:end_comp}. Compared to our base+MA model, our proposed strategy enables our method to transfer component shapes and preserve the texture details missed in generated images from ADGAN and PISE.

\begin{figure}[t!]
   \centering 
\begin{subfigure}{\textwidth}
   \centering 
        \includegraphics[width=0.22\linewidth]{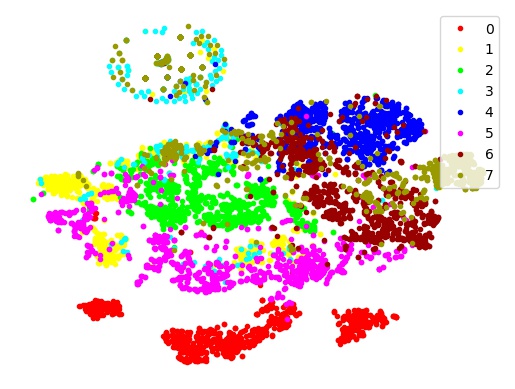}
    \hspace{2pt} 
        \includegraphics[width=0.22\linewidth]{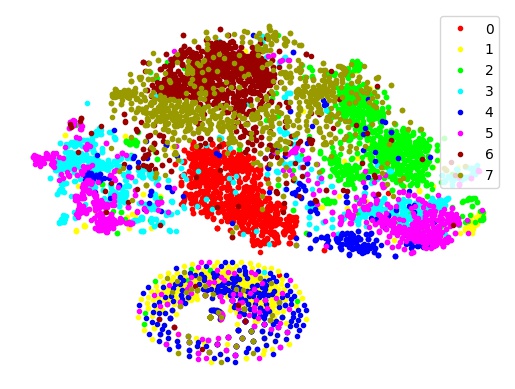}
    \hspace{2pt}     
        \includegraphics[width=0.22\linewidth]{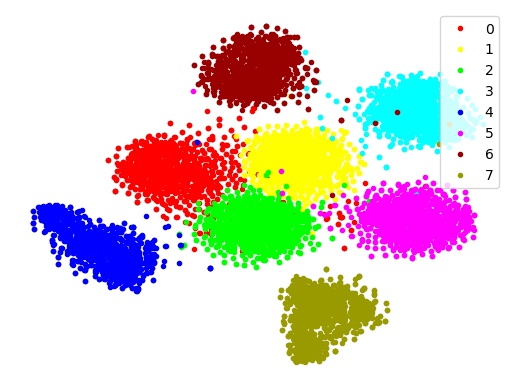}
    \hspace{2pt}
        \includegraphics[width=0.22\linewidth]{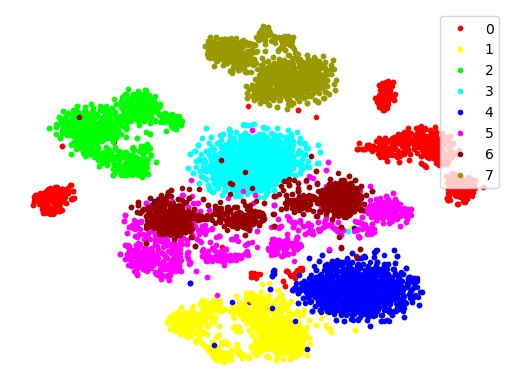}\\
 \vspace{-0.1cm}          
{ \hspace{0pt}  ADGAN \hspace{ 30pt} PISE  \hspace{ 25pt} Base model\hspace{ 20pt}  Ours}      
\end{subfigure} 
   \caption{\label{flowers} Latent code distribution comparison using t-SNE.}
 \label{fig:latent}  
 \vspace{-0.3cm}
\end{figure}

\begin{figure}[t!]
\vspace{-0.35cm}
\captionsetup[subfigure]{justification=raggedright, singlelinecheck=false, labelformat=empty}    
 \centering
    \begin{subfigure}{\textwidth}
  \caption[l]{ \hspace{ 0pt}  Source 1 \hspace{ 170pt}  Source 2}
    \end{subfigure}
\vspace{-4pt}

\includegraphics[width=0.104\textwidth]{posecomp/1_4}
\hspace{-4.5pt}
\includegraphics[width=0.104\textwidth]{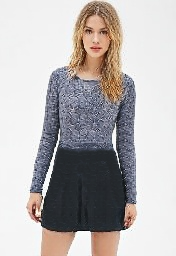}
\hspace{-4.5pt}
\includegraphics[width=0.104\textwidth]{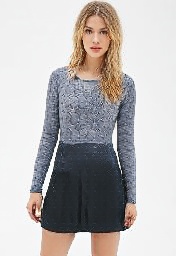}
\hspace{-4.5pt}
\includegraphics[width=0.104\textwidth]{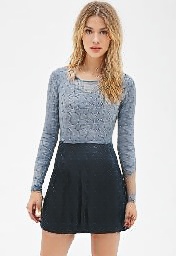}
\hspace{-4.5pt}
\includegraphics[width=0.104\textwidth]{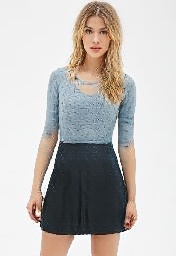}
\hspace{-4.5pt}
\includegraphics[width=0.104\textwidth]{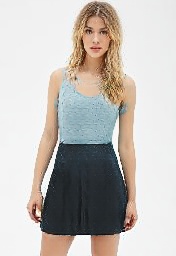}
\hspace{-4.5pt}
\includegraphics[width=0.104\textwidth]{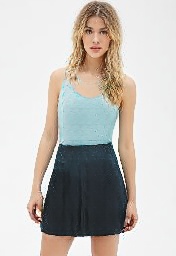}
\hspace{-4.5pt}
\includegraphics[width=0.104\textwidth]{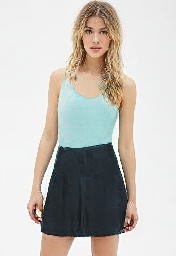}
\hspace{-4.5pt}
\includegraphics[width=0.104\textwidth]{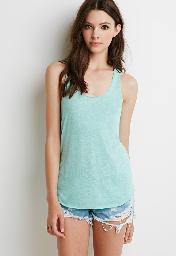}
\caption{Cloth Style Interpolation between two source images.}
\label{fig:interpolation}
\end{figure}

\begin{figure}[t!]
\centering
\includegraphics[width=0.8\textwidth]{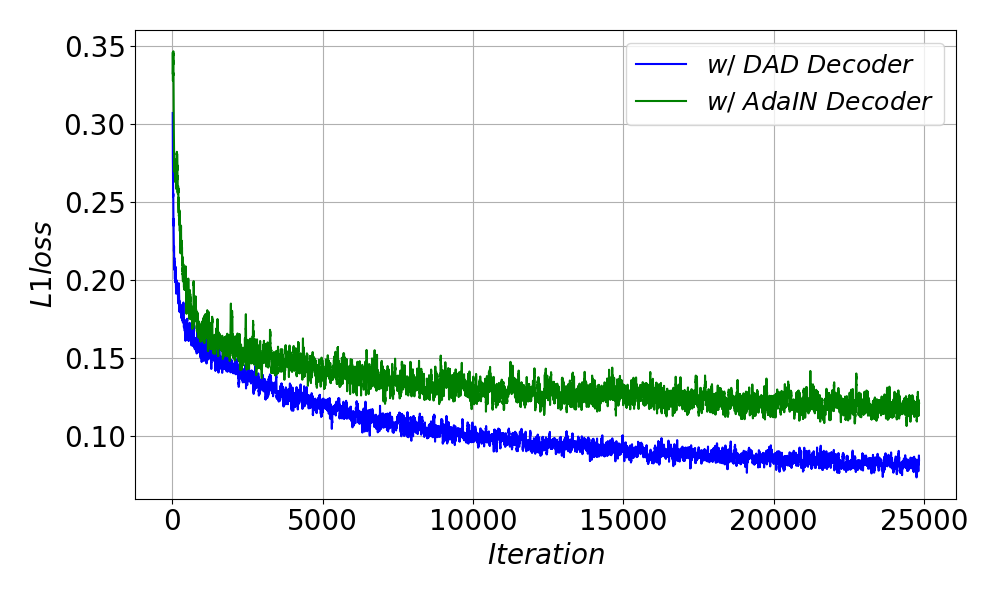} 
\vspace{-0.3cm}
\caption{Comparison between different attribute decoders in terms of loss curve. }\label{fig:losscurve}
\end{figure}

\begin{figure}[t!]
\captionsetup[subfigure]{justification=raggedright, singlelinecheck=false, labelformat=empty}  
\vspace{-0.1cm}
\centering
  \begin{subfigure}[t]{\linewidth}
  \centering
  \includegraphics[height=0.18\textwidth]{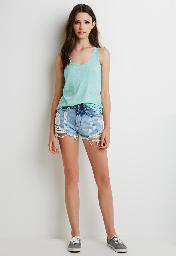} 
    \hspace{-4.5pt}  
  \includegraphics[height=0.18\textwidth]{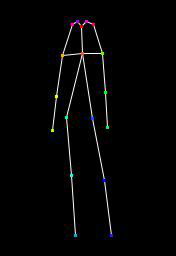} 
    \hspace{-4.5pt} 
  \includegraphics[height=0.18\textwidth]{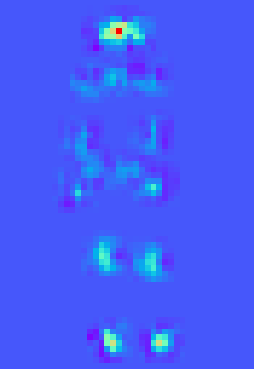} 
    \hspace{-4.5pt} 
  \includegraphics[height=0.18\textwidth]{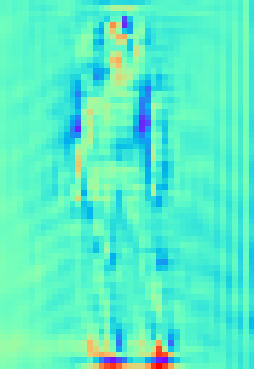} 
    \hspace{-4.5pt}  
  \includegraphics[height=0.18\textwidth]{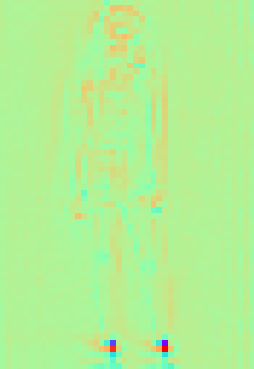}  
    \hspace{-4.5pt} 
  \includegraphics[height=0.18\textwidth]{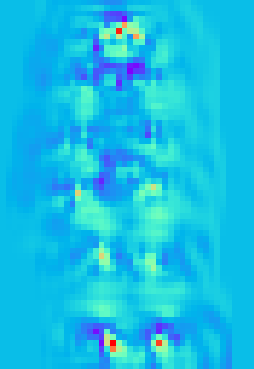}
    \hspace{-4.5pt} 
  \includegraphics[height=0.18\textwidth]{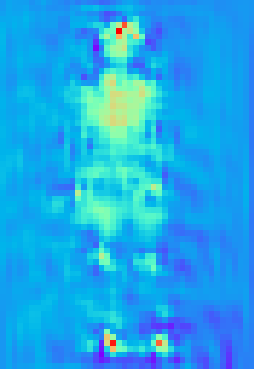}
  \end{subfigure}

 \vspace{-0.36cm} 
     \begin{subfigure}[t]{\linewidth}
     \caption[l]{  \hspace{ 76pt} ResBlk$_0$ \hspace{ 94pt} ResBlk$_8$  }
  \end{subfigure} 
  
  \vspace{2pt}
  
    \begin{subfigure}{\linewidth}
    \centering
  \includegraphics[height=0.18\textwidth]{rebuttal_vis/outt} 
    \hspace{-4.5pt} 
  \includegraphics[height=0.18\textwidth]{rebuttal_vis/outp} 
    \hspace{-4.5pt} 
  \includegraphics[height=0.18\textwidth]{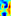} 
    \hspace{-4.5pt}  
  \includegraphics[height=0.18\textwidth]{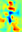} 
    \hspace{-4.5pt} 
  \includegraphics[height=0.18\textwidth]{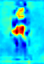} 
    \hspace{-4.5pt} 
 \includegraphics[height=0.18\textwidth]{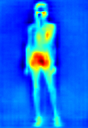} 
    \hspace{-4.5pt}  
  \includegraphics[height=0.18\textwidth]{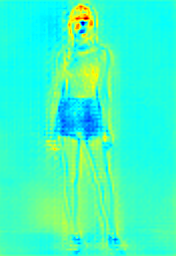} 
    \end{subfigure}

 \vspace{-0.16cm} 
   \begin{subfigure}[t]{\linewidth}
     \caption[l]{  \hspace{ 84pt} $h^2$ \hspace{ 120pt} $h^6$ }
 \vspace{-0.18cm}     
  \end{subfigure} 
  \caption{\label{featmap} Left: input person image and pose; Top: feature maps in Residual Blocks of ADGAN; Bottom: feature maps in Attribute Decoder of DRL-CPG.}  
\vspace{-0.1cm}
\end{figure}

\begin{figure*}[t!]
\captionsetup[subfigure]{justification=raggedright, singlelinecheck=false, labelformat=empty}    
 \centering

 \raisebox{0.03\textwidth }{\makebox[0.12\textwidth]{\parbox{0.01\linewidth}{}}}
 \hspace{-4pt}
 \raisebox{0.03\textwidth }{\makebox[0.12\textwidth]{\parbox{0.01\linewidth}{}}}
 \hspace{-4pt}
 \raisebox{0.03\textwidth }{\makebox[0.12\textwidth]{\parbox{0.01\linewidth}{}}}
 \hspace{-4pt}
\raisebox{0.03\textwidth }{\makebox[0.12\textwidth]{\parbox{0.01\linewidth}{}}}
 \hspace{-4pt}
\includegraphics[width=0.12\textwidth]{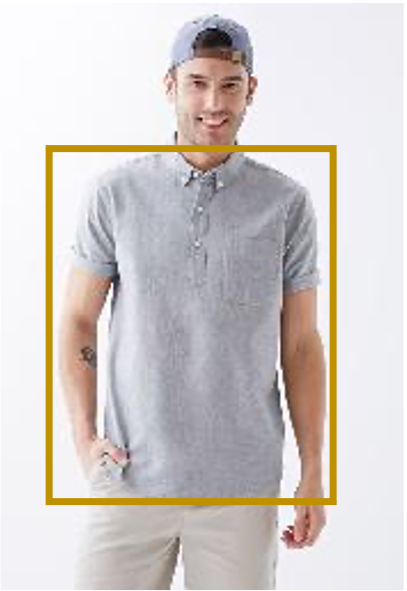}
\hspace{-4pt}
\includegraphics[width=0.12\textwidth]{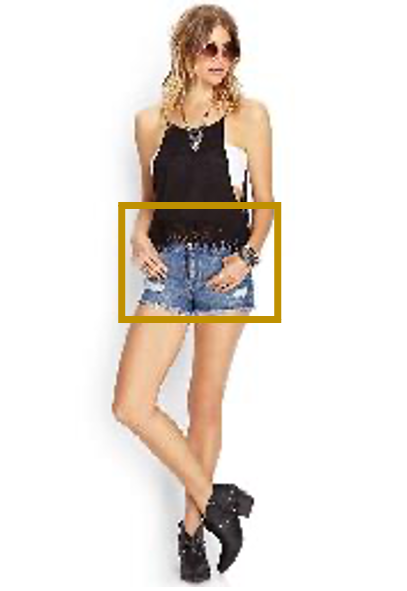}
\hspace{-4pt}
\includegraphics[width=0.12\textwidth]{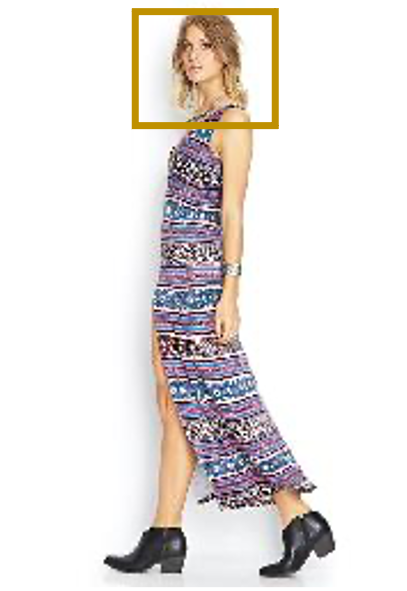}

 \includegraphics[width=0.12\textwidth]{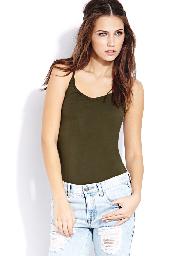}
\hspace{-4pt}  
\includegraphics[width=0.12\textwidth,height=0.174\textwidth]{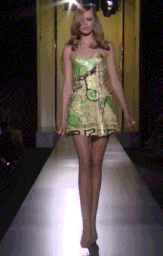}
\hspace{-4pt}
\includegraphics[width=0.12\textwidth]{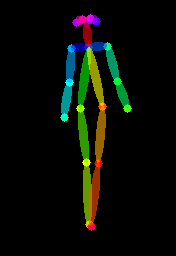}
\hspace{-4pt}
\includegraphics[width=0.12\textwidth]{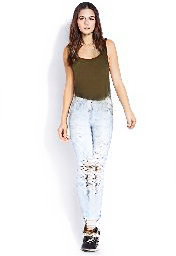}
 \includegraphics[width=0.12\textwidth]{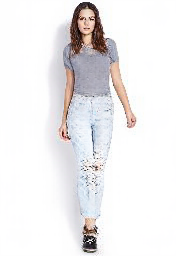}
 \hspace{-4pt}
\includegraphics[width=0.12\textwidth]{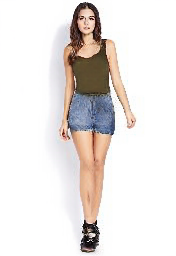}
\hspace{-4pt} 
\includegraphics[width=0.12\textwidth]{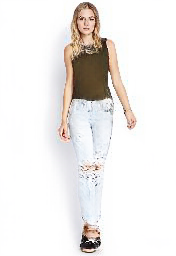}

 \includegraphics[width=0.12\textwidth]{appendixb/1}
\hspace{-4pt}  
\includegraphics[width=0.12\textwidth,height=0.174\textwidth]{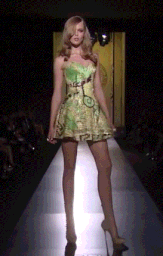}
\hspace{-4pt}
\includegraphics[width=0.12\textwidth]{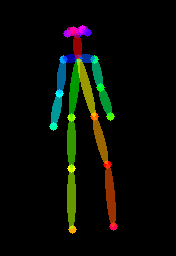}
\hspace{-4pt}
\includegraphics[width=0.12\textwidth]{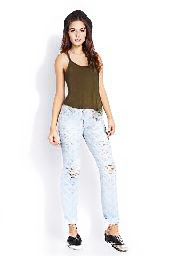}
 \includegraphics[width=0.12\textwidth]{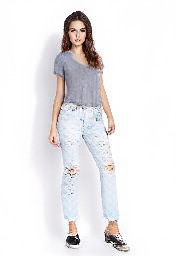}
 \hspace{-4pt}
\includegraphics[width=0.12\textwidth]{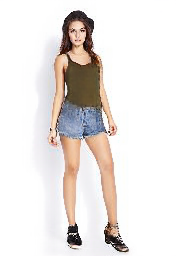}
\hspace{-4pt} 
\includegraphics[width=0.12\textwidth]{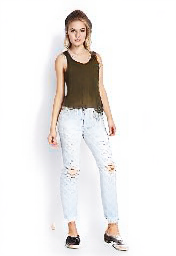}

\begin{subfigure}{\textwidth}
  \caption[l]{ \hspace{ 50pt}  Person   \hspace{ 40pt} Model \hspace{ 30pt} Skeleton  \hspace{ 40pt} Pose \hspace{ 40pt} Cloth  \hspace{ 40pt}Pant \hspace{ 40pt} Head }
    \end{subfigure}
\caption{Video  demonstration  of  Controllable  Person  Synthetic Image  Generation. More detainls are in the supplemental video {\color{blue}  (DRL-CPG$\_$video.mp4)}.}\label{fig:video}
\end{figure*}

\begin{figure}[b!]
\captionsetup[subfigure]{justification=raggedright, singlelinecheck=false, labelformat=empty}    
 \centering

    \begin{subfigure}{\textwidth}
  \caption[l]{  \hspace{ 80pt} Ours\hspace{ 15pt} Ours\hspace{ 12pt} Ours\hspace{ 8pt} ADGAN\hspace{ 6pt} PISE}
    \end{subfigure}

\includegraphics[width=0.134\textwidth]{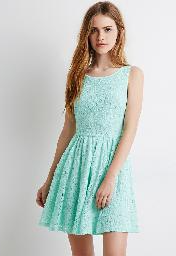}
\hspace{-4.5pt}
\includegraphics[width=0.134\textwidth]{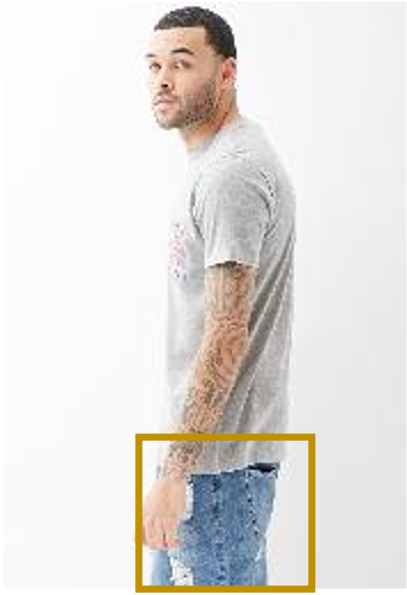}
\hspace{-4.5pt}
\includegraphics[width=0.134\textwidth]{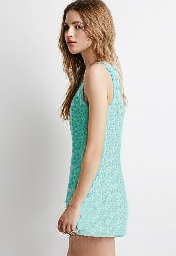}
\hspace{-4.5pt}
\includegraphics[width=0.134\textwidth]{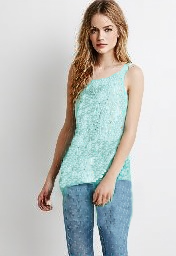}
\hspace{-2.5pt}
\includegraphics[width=0.134\textwidth]{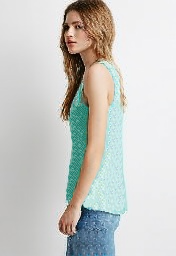}
\hspace{-4.5pt}
\includegraphics[width=0.134\textwidth]{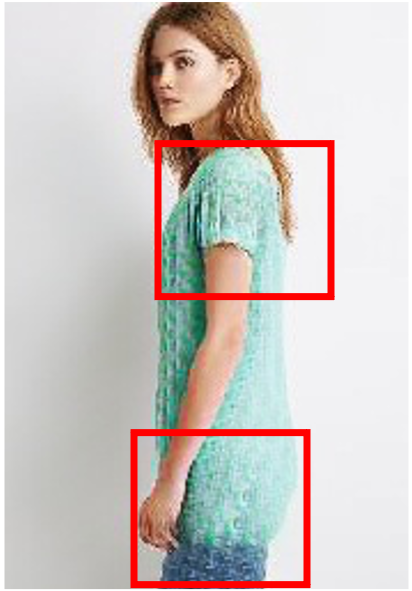}
\hspace{-4.5pt}
\includegraphics[width=0.134\textwidth]{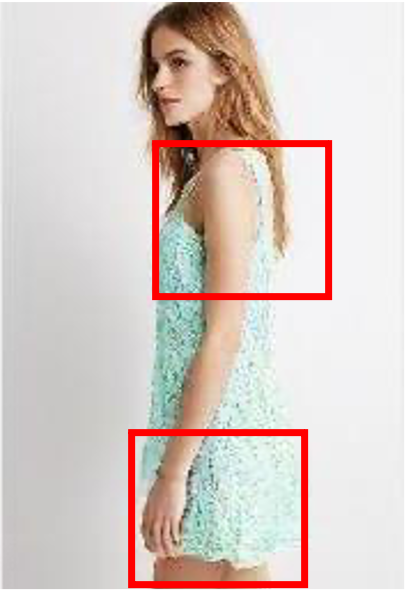}

\includegraphics[width=0.134\textwidth]{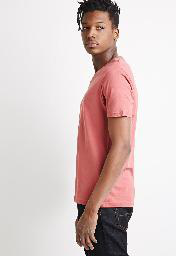}
\hspace{-4.5pt}
\includegraphics[width=0.134\textwidth]{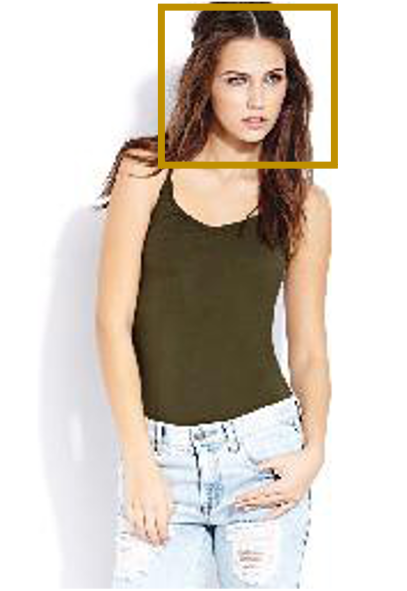}
\hspace{-4.5pt}
\includegraphics[width=0.134\textwidth]{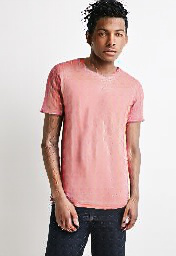}
\hspace{-4.5pt}
\includegraphics[width=0.134\textwidth]{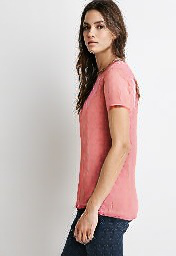}
\hspace{-2.5pt}
\includegraphics[width=0.134\textwidth]{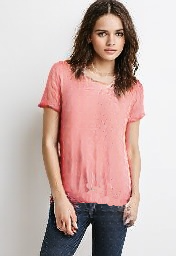}
\hspace{-4.5pt}
\includegraphics[width=0.134\textwidth]{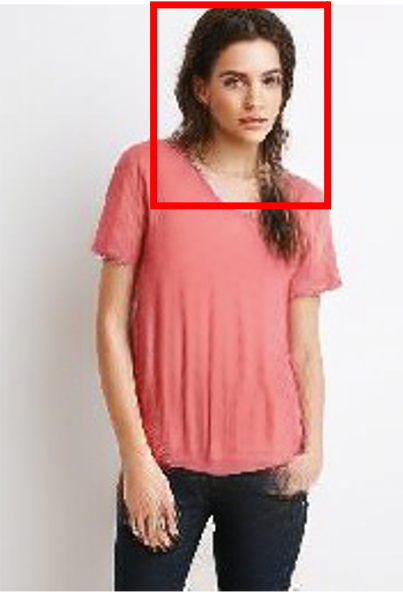}
\hspace{-4.5pt}
\includegraphics[width=0.134\textwidth]{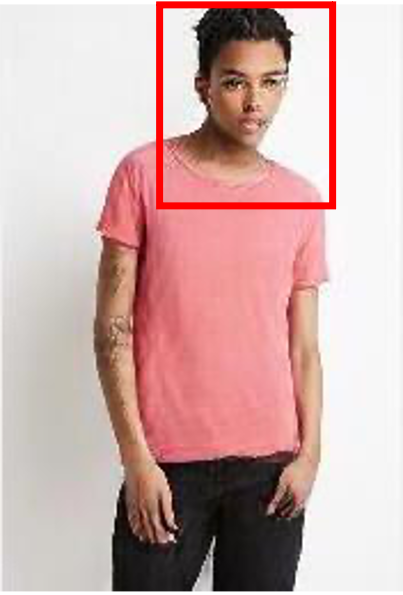}

\includegraphics[width=0.134\textwidth]{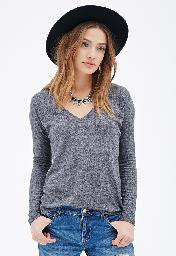}
\hspace{-4.5pt}
\includegraphics[width=0.134\textwidth]{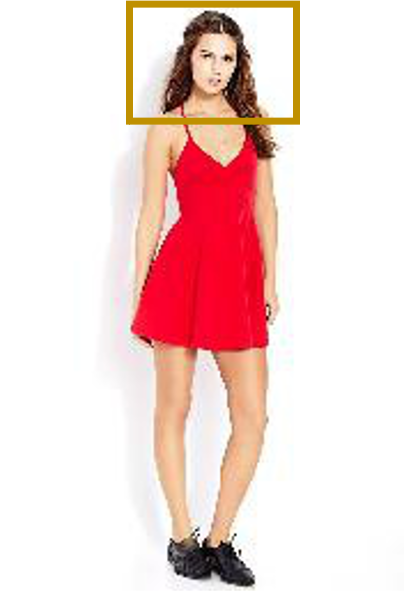}
\hspace{-4.5pt}
\includegraphics[width=0.134\textwidth]{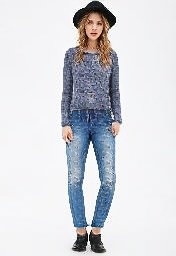}
\hspace{-4.5pt}
\includegraphics[width=0.134\textwidth]{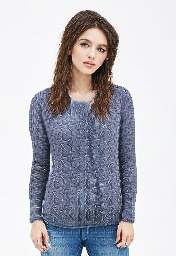}
\hspace{-2.5pt}
\includegraphics[width=0.134\textwidth]{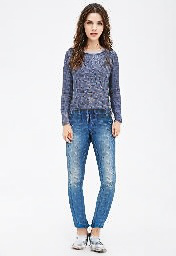}
\hspace{-4.5pt}
\includegraphics[width=0.134\textwidth]{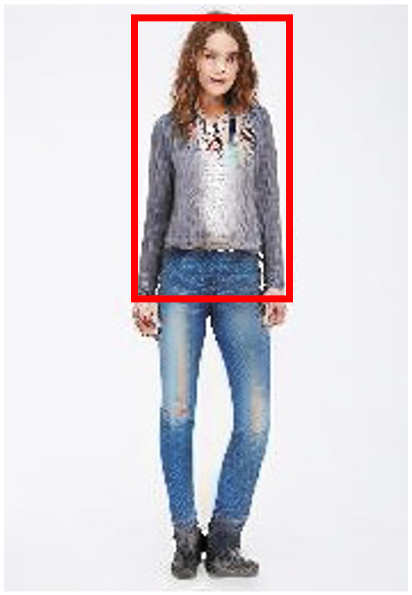}
\hspace{-4.5pt}
\includegraphics[width=0.134\textwidth]{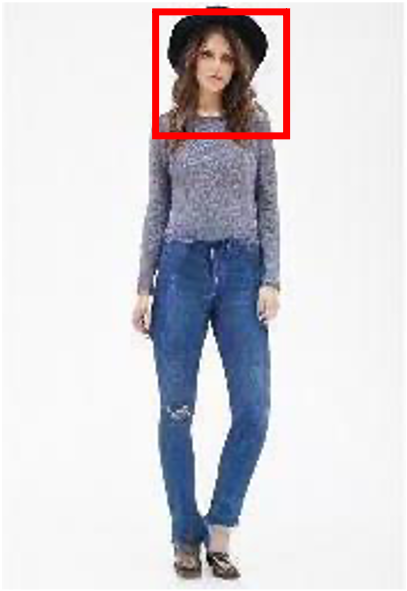}

    \begin{subfigure}{\textwidth}
  \caption[l]{ \hspace{ 4pt}  Person \hspace{ 8pt} Source  \hspace{ 8pt} Pose\hspace{ 13pt} Style\hspace{ 34pt} Pose + style}
    \end{subfigure}
\vspace{-12pt}

\caption{Performance comparison in terms of different tasks.}
\label{fig:pose_cross_add}
\end{figure}

\bfsection{Effectiveness of transformer encoder}
We provide an ablation study on different model sizes of DRL-CPG. In particular, we investigate three different transformer configurations, the ``{\em Small-TX}", ``{\em Medium-TX}" and ``{\em Large-TX}" models. For the ``{\em Small-TX}" model, the embedding dim,  number of layers and MLP ratio are set to be 128, 1, 1, respectively. While those hyperparameters for the ``{\em Medium-TX}" model are 256, 4, 2 and those for the ``{\em Large-TX}" model are 768, 8, 4. ``{\em No-TX}" indicates a CNN attribute encoder without transformers.
As shown in Table \ref{tab:transformer_different_size}, we observe that with the increase of the transformer size, the performance first becomes better and then turns worse. We conclude that the medium model results in a better performance. 

We further apply the trained attribute encoder in our DRL-CPG, Base model, and Base + MA + CL model to a source person image, respectively. As we can observe in Figure~\ref{fig:dec_att}, our attribute encoder with transformers is robust in extracting main attributes from different components, regardless of whether a semantic mask is provided or not. However, the attribute encoder in $\text{Base model}$ can not obtain a proper attention map on different components. This model variant fails to operate controllable person synthetic image generation without a semantic mask.
Again, this observation strongly demonstrates the efficacy of our well-designed DRL-CPG. As we can observe in Figure~\ref{fig:dec_att}, our attribute encoder with transformers is robust in extracting main attributes from different components, regardless of whether a semantic mask is provided or not.   

\bfsection{Effectiveness of disentangling latent representation}
To illustrate the encoded latent space learned by different model variants, we show the feature distribution comparison in Figure \ref{fig:latent} using t-SNE. With semantic masks available, our encoded features can form more compact and separable clusters than the ADGAN and PISE networks. Thanks to the transformer encoder, our base model trained without any strategy can project all the intermediate features to well-separated regions. However, these latent features do not learn semantic meanings as proven by the results listed in the third column of Figure \ref{fig:Ablation}. 
Therefore, although the distribution of our base model looks well separated, it cannot deal with some hard or ambiguous cases, while our DRL-CPG can handle them well. In Figure \ref{fig:interpolation}, we demonstrate the interpolation results. The smooth transition between two clothe styles proves our model learns a well latent structure. 

\bfsection{Effectiveness of Attribute Decoder}
To evaluate the effectiveness of our proposed DAD-based attribute decoder, we train a variant of our model, which consists of a decoder used by ADGAN. As AdaIN residual block is the main building block of ADGAN's decoder, we denote this decoder as AdaIN decoder. Specifically, we take the same encoder to obtain the component attribute representations and feed them to different decoders to synthesize person images. The loss curve for the effects of our DAD-based attribute decoder during training is shown in Figure \ref{fig:losscurve}. We observe that our decoder achieves lower L1 loss. Our DAD module integrating semantic representations into different spatial regions outperforms the AdaIN which only focuses on the semantic representations and ignores the spatial information. Thus, our proposed DRL-CPG can synthesis better person images.

Additionally, we compare the feature maps from the residual blocks of ADGAN and the attribute decoder of our model in Figure \ref{featmap}. These feature maps show that our decoder establishes the relation between different joints and gradually constructs the person under guidance, while ADGAN focuses on local regions around joints.
and creates unsmooth features.

\section{More Visualization Results}

{\noindent\bf Controllable Person Synthetic Image Generation.}
For the original person image, our proposed DRL-CPG can change its component attributes ({\em e.g.}, upper clothes, pants, and
head) with another person image providing the desired attribute. In Figure \ref{fig:pose_cross_add}, we show more results on the challenging task, which requires to transfer multiple attributes into the source person. By learning the disentangled representations, our model successfully transfers multiple attributes into the person, while other baseline methods struggle with reconstructing the attributes. 

{\noindent\bf Video demonstration of Controllable Person Synthetic Image Generation.}
To further demonstrate the capability of our model in learning disentangled representation, we perform video generation. Given a source person, a video containing a walking motion of a fashion model and desired attributes, \eg, cloth, pant, hair, our model can synthesize a motion for the source
person under a series of target poses extracted from the fashion model, and simultaneously sharing component attributes transferred from other person. In Figure \ref{fig:video} we show the synthetic person with different attributes, and list two typical poses extracted from the fashion model as reference. It is strongly recommended to watch the supplemental video {\color{blue}  (DRL-CPG$\_$video.mp4)} for the
visualization, which proves the effectiveness of controllable person synthetic image generation. This demonstrates that our model learns a smooth and well-distributed latent space that is constituted of various human attributes of the person images, including pose, upper clothes, pants, head and so on.

\section{Conclusion}
In this paper, we have proposed a novel framework with transformers to learn disentangled representation with transformers for controllable person image synthesis. Our model learns well disentangled latent representations via the proposed random component mask-agnostic strategies and is able to operate on human editing. While very promising results have been achieved in all experiments, the texture details of the synthesized images are not entirely realistic. We plan to synthesize images in high-resolution and improve the quality of our future work.

%
%
\ifCLASSOPTIONcaptionsoff
  \newpage
\fi

{\small
\bibliographystyle{IEEEtran}
\bibliography{egbib}
}

\end{document}